\documentclass[nohyperref]{article}

\usepackage{microtype}
\usepackage{graphicx}
\usepackage{subfigure}
\usepackage{booktabs} 
\usepackage{multirow}
\usepackage{bbm}
\usepackage{hyperref}

\usepackage[accepted]{icml2022}

\usepackage{amsmath}
\usepackage{amssymb}
\usepackage{mathtools}
\usepackage{amsthm}
\usepackage{makecell}
\usepackage{listings}
\usepackage{xcolor}

\definecolor{codegreen}{rgb}{0,0.6,0}
\definecolor{codegray}{rgb}{0.5,0.5,0.5}
\definecolor{codepurple}{rgb}{0.58,0,0.82}
\definecolor{backcolour}{rgb}{1,1,1}

\lstdefinestyle{mystyle}{
    backgroundcolor=\color{backcolour},   
    commentstyle=\color{codegreen},
    keywordstyle=\color{magenta},
    numberstyle=\tiny\color{codegray},
    stringstyle=\color{codepurple},
    basicstyle=\ttfamily\footnotesize,
    breakatwhitespace=false,         
    breaklines=true,                 
    captionpos=b,                    
    keepspaces=true,                 
    numbers=none,                    
    numbersep=5pt,                  
    showspaces=false,                
    showstringspaces=false,
    showtabs=false,                  
    tabsize=2,
    linewidth=22cm
}

\usepackage[capitalize,noabbrev]{cleveref}

\theoremstyle{plain}

\theoremstyle{definition}

\theoremstyle{remark}

\usepackage[textsize=tiny]{todonotes}

\icmltitlerunning{Mold into a Graph: Efficient Bayesian Optimization over Mixed-Spaces}

\begin{document}

\twocolumn[
\icmltitle{Mold into a Graph: Efficient Bayesian Optimization over Mixed-Spaces}



\icmlsetsymbol{equal}{*}

\begin{icmlauthorlist}
\icmlauthor{Jaeyeon Ahn}{yyy}
\icmlauthor{Taehyeon Kim}{yyy}
\icmlauthor{Se-young Yun}{yyy}
\end{icmlauthorlist}

\icmlaffiliation{yyy}{Kim JaeChul Graduate School of AI, KAIST, Seoul, South Korea}

\icmlcorrespondingauthor{Jaeyeon Ahn}{dkswodus49@kaist.ac.kr}

\icmlkeywords{Machine Learning, ICML}

\vskip 0.3in
]



\printAffiliationsAndNotice{}  

\begin{abstract}
Real-world optimization problems are generally not just black-box problems, but also involve mixed types of inputs in which discrete and continuous variables coexist. Such mixed-space optimization possesses the primary challenge of modeling complex interactions between the inputs. In this work, we propose a novel yet simple approach that entails exploiting the \textit{graph} data structure to model the underlying relationship between variables, i.e., variables as nodes and interactions defined by edges. Then, a variational graph autoencoder is used to naturally take the interactions into account. We first provide empirical evidence of the existence of such graph structures and then suggest a joint framework of graph structure learning and latent space optimization to adaptively search for optimal graph connectivity. Experimental results demonstrate that our method shows remarkable performance, exceeding the existing approaches with significant computational efficiency for a number of synthetic and real-world tasks.
\end{abstract}

\vspace{-20pt}
\section{Introduction}
\label{introduction}
Black-box problems\,\cite{audet2017derivative, Bajaj2021} are prevalent in real-world applications where the objective formulation is unknown, and the only accessible information is the evaluation of solution candidates. Such optimization is typically accompanied by further challenges such as expensive evaluation costs\,\cite{jones1998expensivebbo, wang2004mode}, high-dimensional search spaces\,\cite{shan2010survey, mei2016competitive}, or a mixture of input data types\,\cite{hutter2011sequential}. While a number of studies have been conducted on the first two challenges, the problem of mixed input spaces, i.e., a mixture of discrete (either nominal or ordinal) and continuous inputs, remains relatively under-explored despite its widespread presence. For instance, to design a space rocket, it is necessary to determine what type of fuel to use (nominal variable), how many nozzles to use (ordinal variable), and the volume of propellant to use (continuous variable). Acquiring the performance of a specific configuration for the task is a huge cost burden. However, more crucially, without domain-specific expertise, it is difficult to determine how the variables interact, especially when the variables have no numerical ordering (e.g., types of fuel contains $\mathtt{\{liquid\ hydrogen, aluminum, \dots}\}$), which often expands to a large number of possible values as well. 

Bayesian Optimization\,(BO)\,\cite{snoek2012practical, shahriari2016, frazier2018tutorial, turner2021bayesian} is a predominant method for solving black-box functions. Its sample efficiency and wide applicability derived from an expressive surrogate model make it stand out among different algorithms. The popular choice for the surrogate model is the Gaussian Process\,(GP)\,\cite{Rasmussen2004} due to its capability of expressing a broad range of statistical models which it gains from the properties of normal distribution and means of kernels. Regarding its strength, a few recent GP-based BO works have arisen to explicitly address the mixed-space optimization problems and suggest different ways of modeling GP. Nevertheless, existing works have clear limitations, as the mechanism for interaction learning either lacks its guarantee of expressiveness due to the kernel being manually designed\,\cite{ru2020bayesian, wan2021think} or requires a substantial computational cost to train additive GP\,\cite{deshwal2021bayesian}.

\begin{figure}[t]
\begin{center}
\centerline{\includegraphics[width=\columnwidth]{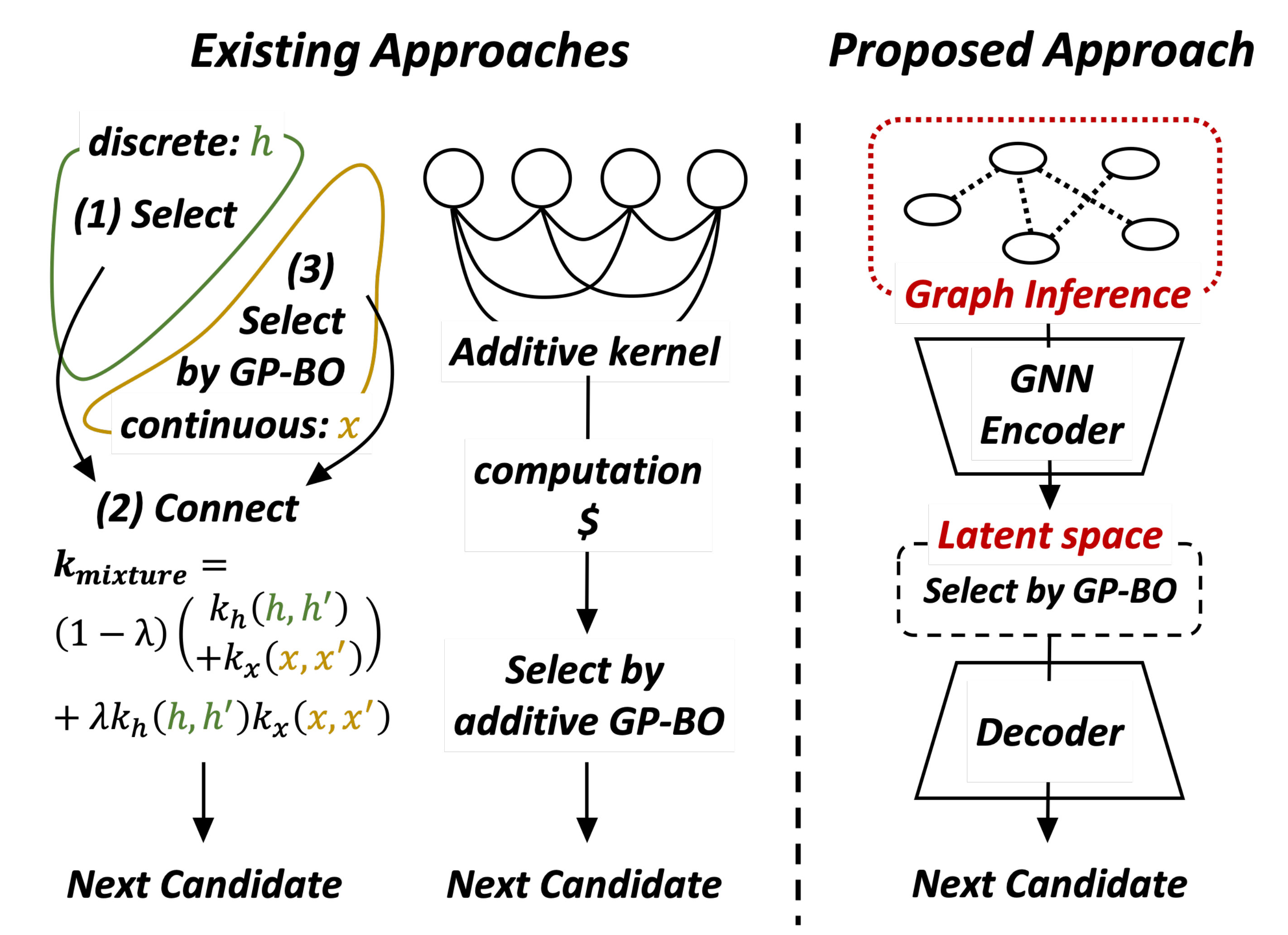}}
\vspace{-10pt}
\caption{Comparison of the proposed algorithm with existing approaches. The left figure shows previous approaches of separate modeling for different data types\,\cite{ru2020bayesian, wan2021think} whereas the middle figure shows the adoption of additive GP\,\cite{deshwal2021bayesian}. Our algorithm conducts graph inference in the given search space and proceeds latent space optimization, thereby preserving a unified search space while also achieving search efficiency.}
\label{overview}
\end{center}
\vskip -0.2in
\vspace{-20pt}
\end{figure}

To resolve such issues, we propose a novel framework, termed GEBO\,(\underline{\textbf{G}}raph-molding for \underline{\textbf{E}}fficient \underline{\textbf{B}}ayesian \underline{\textbf{O}}ptimization over mixed-spaces), which is composed of latent space optimization and graph structure learning that jointly operate in an end-to-end fashion. The key idea of GEBO is to simply mold the given data into a \textit{graph}, i.e., each variable corresponds to an individual node and edges define the interactions between them, and learn a continuous latent space with a variational graph autoencoder\,(VGAE) (\autoref{overview}). Thus, we can not only naturally integrate relational information into the latent graph embedding but also take full advantage of GP-BO within the learned space. 

To verify our approach, we empirically demonstrate the existence of suitable graph structures that successfully facilitate the optimization by conducting an exhaustive search of possible connectivities. For simplicity, we search over undirected graphs. Then, we provide numerical evidence of the presence of certain nodes (i.e., input variables) that have a notable correlation between their importance within a graph and the corresponding graph's suitability. Based on this observation, we introduce a graph learning mechanism that consummates our GEBO approach, coined as \textit{nested} MAB, which effectively generates an adapted graph structure for interaction modeling. Our key contributions are as follows:

\vspace{-12pt}
\begin{itemize}
\item We suggest a novel graph-based approach for mixed-space optimization in which we simply mold the data into a \textit{graph} to model the complex interactions. We introduce a joint framework of graph structure learning and VGAE-based optimization so that we can attain not only interaction learning without human intervention but also computational efficiency and scalability.
\vspace{-7pt}
\item We empirically show the existence of specific nodes (inputs) for a given task in which, when centered in the graph, i.e., when they are relatively more influential than other nodes, the optimization tends to be superior. Based on this observation, we develop an adapted graph learning mechanism referred to as \textit{nested} MAB.
\vspace{-7pt}
\item  Our algorithm demonstrates effective performance on synthetic and diverse domains of real-world tasks, especially showing superior performance in terms of its computational efficiency compared to the state-of-the-art methods, which is further proven in high-dimensional cases (\autoref{finalperf}). The implementation of GEBO is available at \url{https://github.com/jjaeyeon/GEBO.git}.
\end{itemize}
\begin{table}[t]
\caption{Performance comparison for synthetic and various domains of real-world tasks, e.g., scientific environment calibration, mechanical design problems, and control parameter tuning, etc. The results are rated with the time constraint of 60 seconds in wall clock time. The task scale expands from 4 to 53 dimensions, and our algorithm surpasses the state-of-the-art algorithms in all cases.\label{finalperf}}
\vspace{-5pt}
\begin{center}
\begin{small}
\begin{sc}
\resizebox{\columnwidth}{!}{\begin{tabular}{lccccr}
\toprule
Task & CoCaBO & CASMOPOLITAN & HyBO & \textbf{GEBO (Ours)}\\
\midrule
\makecell[l]{Environment\\Calibration} & -0.189$\pm$0.176 & -0.131$\pm$0.074 & -0.195$\pm$0.148 & \textbf{-0.034$\pm$0.028}\\
\midrule
\makecell[l]{Pressure Vessel\\Design}  & -0.026$\pm$0.014 & -0.044$\pm$0.016 & -0.502$\pm$0.224 & \textbf{-0.027$\pm$0.000}\\
\midrule
\makecell[l]{Speed Reducer\\Design}    & -1.178$\pm$0.001 & -1.245$\pm$0.014 & -1.425$\pm$0.000 & \textbf{-1.178$\pm$0.001}\\
\midrule
\makecell[l]{Neural Network\\HPO}    & 0.926$\pm$0.004 & 0.930$\pm$0.004 & 0.882$\pm$0.000 & \textbf{0.934$\pm$0.003}\\
\midrule
\makecell[l]{Robot Pushing\\Control}   & 3.331$\pm$0.384 & 3.094$\pm$0.042 & 3.179$\pm$0.174 & \textbf{4.138$\pm$0.196}\\
\midrule
\makecell[l]{Ackley53C}                & -1.238$\pm$0.004 & -1.251$\pm$0.008 & -1.260$\pm$0.000 & \textbf{-1.231$\pm$0.005}\\
\bottomrule
\end{tabular}}
\end{sc}
\end{small}
\end{center}
\vskip -0.1in
\vspace{-17pt}
\end{table}

\vspace{-15pt}
\section{Preliminaries}
\label{preliminaries}

In this paper, we consider the black-box optimization problems over mixed-spaces in which the inputs contain a mixture of discrete and continuous variables. Different names are often used to refer to the noncontinuous data types in machine learning literature. Here, we use "\textit{discrete}" variables to collectively call the variables that are not continuous, which can be further categorized into two: \textit{nominal} variables (values by names) and \textit{ordinal} variables (values with numerical order). We do not discriminate between nominal and ordinal variables because our approach is applicable to both type.

Let $\mathcal{X}$ be a mixed-space where element $\mathbf{x} = [\mathbf{x_{d}}, \mathbf{x_{c}}]$, $\forall{\mathbf{x}}\in\mathcal{X}$ consists of $k_{d}$ number of discrete variables $\mathbf{{x}_{d}}$ and $k_{c}$ number of continuous variables $\mathbf{{x}_{c}}$. The discrete variables may have multiple possible values, i.e. $x_{d_{i}}$ may have $n_{i}$ number of possible values. Our proposed approach, GEBO, reformulates the data into a \textit{graph} where each dimension matches to individual node, i.e. ($k_{d} + k_{c}$) number of nodes total, and the nodes are linked by certain connectivity $A$. Consequently, for $|\mathcal{X}|=S$, we obtain $S$ number of graphs, each with same connectivity $A$ but a different set of node representations. For node representations, the discrete variables are transformed into one-hot encoding (i.e. $x_{d_{i}}$ is converted into a $n_{i}$-dimensional vector) while continuous variables are unit-scaled by the given bounds, and then projected into shared node feature space through individual projection layers. 

\vspace{-10pt}
\section{Related Work}
\label{relatedwork}
\subsection{BO for Mixed Spaces}
\vspace{-5pt}
\paragraph{Non-GP based BO.} Early BO methods that adopted a non-GP surrogate model naturally appear to be compatible for mixed-space optimization. SMAC\,\cite{hutter2011sequential} uses random forests\,(RFs)\,\cite{breiman2001random} as its surrogate model. However, the prediction distribution is relatively less accurate due to its innate randomness and the algorithm is also prone to overfitting. Another method that uses a tree-based surrogate model is TPE\,\cite{bergstra2011algorithms}. It uses nonparameteric Parzen kernel density estimators to model the distribution of each dimension independently, which limits not only its scalability but also its  expressivity because of the absence of interdependency consideration. 

\vspace{-15pt}
\paragraph{GP-based BO.} Recent GP-based BO methods surpass all the aforementioned algorithms. CoCaBO\,\cite{ru2020bayesian} is the first work that explicitly tackles the mixed-space problems. CoCaBO operates with separate search strategies for different data types; it first selects discrete variables with a multi-armed bandit\,(MAB) and then optimizes a continuous subspace with GP-BO. To connect the separated subspaces, it suggests the use of a specific kernel to establish GP, which computes a mixture of the sum and product of a Hamming kernel (categorical) and a radial basis function (RBF) kernel (continuous). As a result, the hand-crafted kernel has restrictions on learning distinct interactions between different pairs of variables. In addition, the MAB mechanism makes the algorithm impractical for the problems with high-dimensional discrete variables. Specifically targeting this limitation in scalability, CASMOPOLITAN\,\cite{wan2021think} expands applicability to high-dimensional problems by adopting the local search method, which recently has been getting attention due to its efficiency in BO literature\,\cite{eriksson2020scalable}. Instead of selecting discrete variables with MAB, it conducts a local search within each subspace where the discrete subspace is constructed by a combinatorial graph containing all possible configurations of the variables as nodes. It then uses the same kernel as that in CoCaBO to reconnect the broken spaces. Consequently, the algorithm inherits the shortcoming of the limited kernel expressiveness. HyBO\,\cite{deshwal2021bayesian} introduces a diffusion kernel that can be naturallly applied to both discrete and continuous inputs. With the diffusion kernel as a base kernel for each dimension, HyBO builds up an additive kernel to learn different orders of interactions between the inputs. This additive GP formulation\,\cite{duvenaud2011additive} is widely known for its critical drawback of requiring a large computational cost to train the model to attain sufficient predictability. We therefore propose a new perspective that eliminates all the abovementioned limitations above by modeling the interaction with a \textit{graph}, thereby retaining a unified search space and efficiently integrating relational information through the use of graph neural networks (GNNs). 

\vspace{-5pt}
\subsection{Graph Inference}
\vspace{-5pt}
Graphs are a common tool for expressing data relationships (edges) between different entities (nodes) in machine learning literature. Recent advances in deep learning on graph-structured data have shown remarkable performance for various tasks by harnessing the relational information for the target objective\,\cite{wu2021graphsurvey}. Through GNN, the local message passing allows each node to take its neighborhood into account, forming a certain context as a whole. Applying an accurate graph structure to a given task is important to secure satisfactory performance of GNN performance, which has naturally evoked a number of studies on graph structure learning\,\cite{zhu2021deep}. Despite the success of deep graph structure learning, random graphs indeed show surprising performances, especially in the field of neural architecture search \cite{xie2019exploring, you2020graph}. \citet{xie2019exploring} applies classic random graph generation models to neural architecture structures which resulted in competitive results comparing to state-of-the-art models. \citet{you2020graph} introduces a novel graph representation for neural architecture and investigate the relationship between the graph structure and architecture performance. They introduce a modified random graph generator to extensively explore the graph space, which eventually draws high-performing connectivities. Inspired from these works, we utilize random graphs to model the underlying interdependencies. The details are discussed in the method section. 

One of the most popular algorithms used for network analysis is PageRank (PR)\,\cite{page1999}. The key idea of PR is that a page is as important as the pages that are linked to it. Consequently, it assigns importance to each node based on the number of incoming links and the importance of corresponding source nodes. The formulation is as follows: 
\newline
\vspace{-6pt}
\begin{equation}
    PR(n_{i}) = \frac{1-d}{N} + d\sum_{n_{j} \in \mathcal{N}(n_{i})} \frac{PR(n_{j})}{L(n_{j})}
    \vspace{-6pt}
\end{equation}
\newline
where N is the total number of nodes, $\mathcal{N}$ is a set of neighbor nodes with incoming links, $L$ is the number of outgoing links of a node, and $d$ is the damping factor. While PR is commonly applied to directed graphs, it is also often used for undirected graphs \cite{Jinghua, Zeinab, Gabor, Nicola}, in which the connected edges are treated as two separate directed edges pointing in opposite directions. Note that the PR values for an undirected graph are not exactly proportional to the degree distribution of the graph \cite{grolmusz2015}, except for the case when the given structure is a regular graph. We use PR to analyze the correlation between individual nodes (inputs) and the graph's performance.
\begin{figure*}[ht]
\vskip 0.2in
\begin{center}
\centerline{\includegraphics[width=0.8\textwidth]{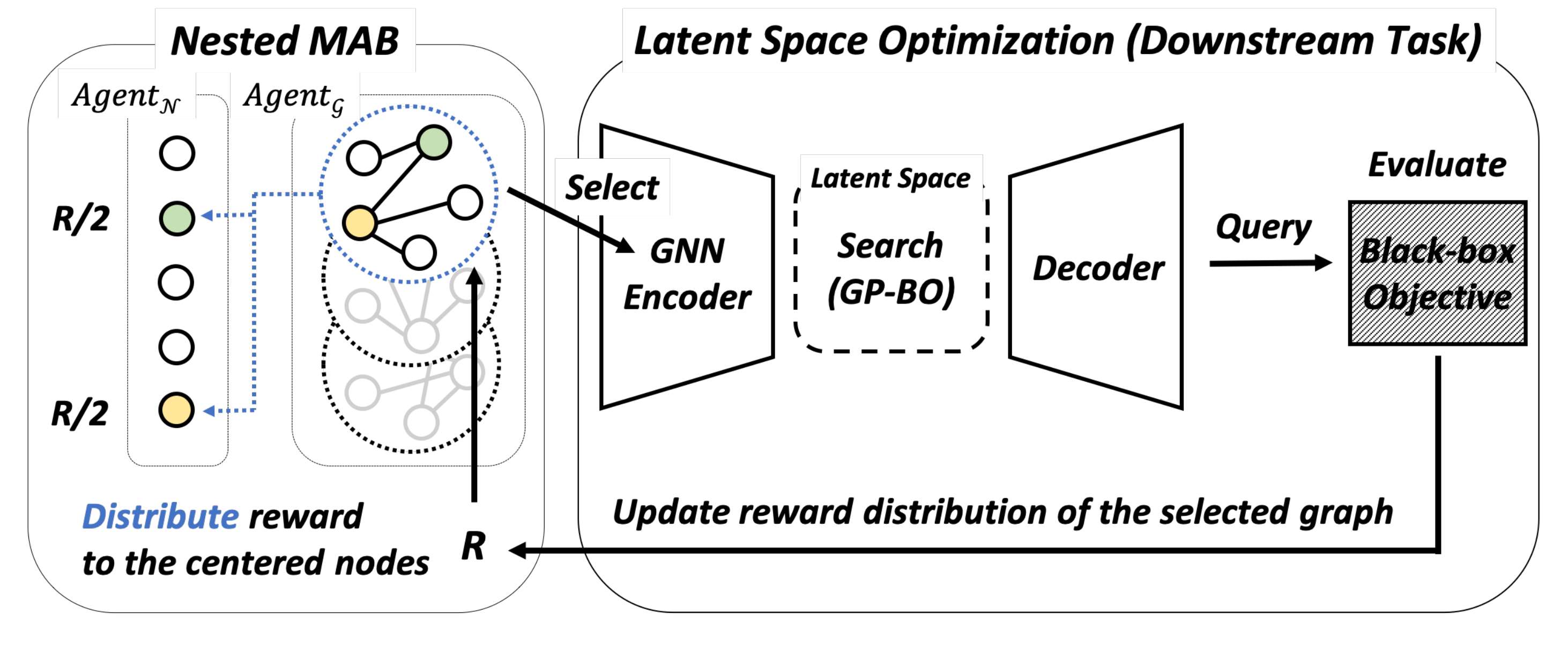}}
\caption{Illustration of GEBO for a 5-dimensional task with $K=3$ candidate graphs as an example. The two modules, \textbf{\textit{nested} MAB for graph learning} and \textbf{VGAE-based LSO}, jointly learn in an end-to-end fashion. For every iteration, the former first selects the randomly centered graph, and then the latter uses the selected graph to encode, conducts optimization with latent graph embedding through GP-BO, and queries the evaluation of the decoded new input. The evaluation is used as a reward $\mathbf{R}$ for both the encoded graph and centered nodes in the corresponding graph. For the node reward, the rewards are divided by the number of centered nodes, e.g., the figure illustrates the case of two centered nodes. With the incremented data, VGAE is retrained before proceeding to the next iteration.}
\label{gebo}
\end{center}
\vskip -0.2in
\vspace{-10pt}
\end{figure*}

\vspace{-5pt}
\section{Method}
\label{method}
\subsection{GEBO: Approach}
\label{approach}
\vspace{-5pt}
Our approach, termed as GEBO, can be decomposed into two parts, latent space optimization (the downstream task) and the graph learning module, which jointly operate throughout the optimization process. An overview of our algorithm is presented in \autoref{gebo}.
\vspace{-10pt}
\paragraph{Latent space optimization\,(LSO).} Learning a latent manifold via a deep generative model is a predominant method in optimizing either structured or high-dimensional spaces\,\cite{siivola2021good} across various fields of applications, such as neural architecture search \cite{elsken2019neural, ren2021comprehensive} and molecular design \cite{sanchez2018molecular, elton2019molecular}. In the embedded spaces, we can make use of the efficiency of conventional black-box algorithms such as GP-BO. In this work, we cast the problem of mixed-space optimization into LSO of a graph-structured space by reformulating the data into a graph. Our work differs from the existing LSO methods in that the target spaces are not originally structured. With the learned graph embedding, GP-BO is conducted to select a new representation and decode it into a valid configuration with the trained generative model. Then, the outcome is used for the next evaluation. To solely test the efficacy of the graph-molding method, the standard choice (i.e. VGAE\,\cite{kipf2016variational} with a multilayer perceptron (MLP) for projection) is adopted for the generative model, and we follow the recent consensus in the training scheme of the LSO framework\,\cite{tripp2020sampleefficient}. We present the pseudocode and include a detailed description of implementation in \autoref{algodescription}. 

\textbf{Graph learning.} To demonstrate the existence of graph structure that enables effective LSO, we first evaluate over all possible graph structures of toy examples (Section\,\ref{proofofconcept}). Regarding the notable pattern observed between specific graph properties and the graph's performance, we suggest a graph structure learning mechanism, named as \textit{nested} MAB, that jointly searches for better structure that reflects informative relationship for given task (Section\,\ref{proposed}). 

\begin{figure*}[ht]
\vskip 0.1in
\begin{center}
\centerline{\includegraphics[width=\textwidth]{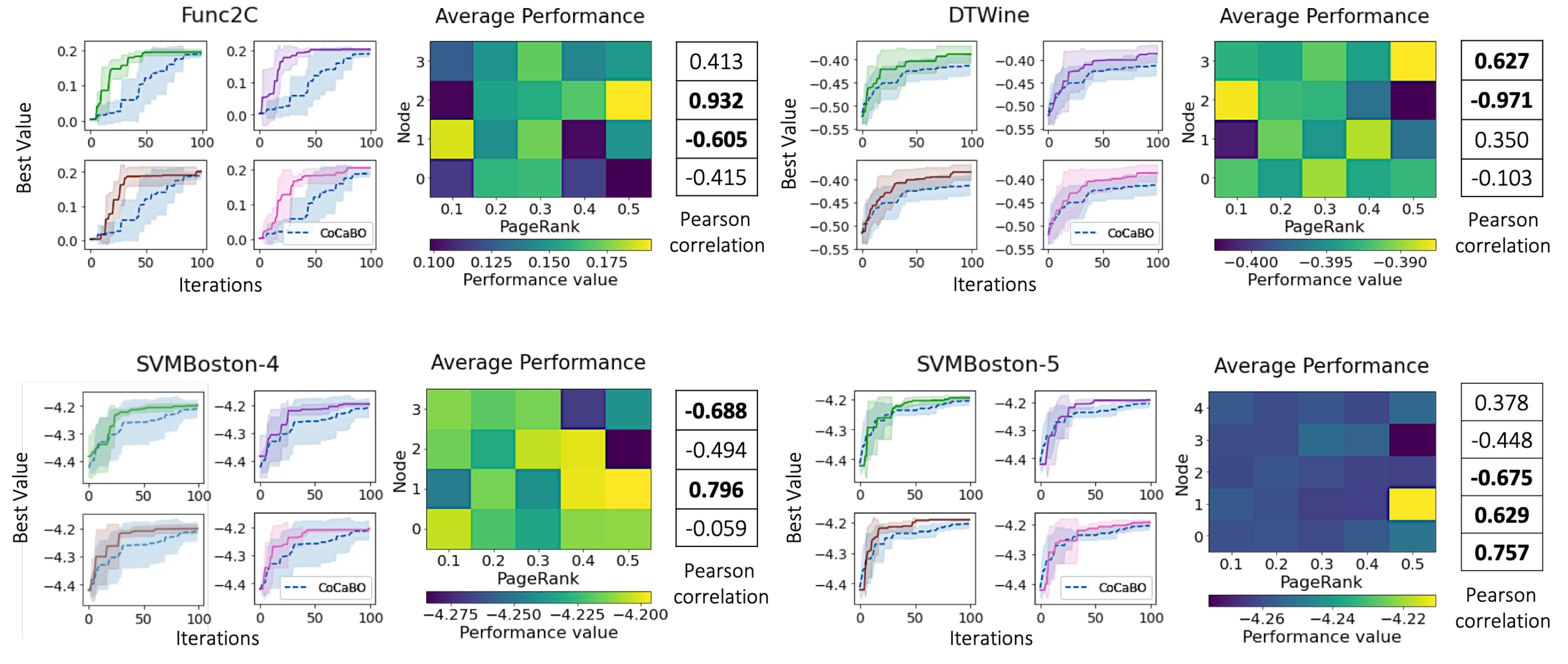}}
\caption{For each task, the results are reported as mean $\pm$ 1 standard deviation, and high performing cases of the graph-molding are plotted using solid colored lines. On the right side of the plots, we visualize the average performance of the graphs by heatmaps that fall into the divided segments of PageRank values for each node (i.e. input; refer to the task descriptions in Section\,\ref{proofofconcept} for node indices of inputs denoted in the $y$-axis). Along with the visualization, we present the Pearson correlation coefficients between nodes' PR values and the corresponding graph's performance where the notable values (absolute values over the threshold of 0.6) are marked in bold.}
\label{toyproof}
\end{center}
\vskip -0.3in
\end{figure*}
\vspace{-6pt}

\subsection{Proof of Concept}
\label{proofofconcept}
\vspace{-5pt}
In this subsection, we aim to verify that there exist optimal graph structures that effectively reflect the informative relationship between variables for target optimization. We conduct an exhaustive search of possible graph structures of four tasks, consisting of one synthetic task and three real-world tasks on hyperparameter optimization of standard machine learning models as follows\footnotemark[1]:
\vspace{-10pt}
\begin{itemize}
    \item \textbf{Func2C}\,\cite{ru2020bayesian} is a synthetic task with 2 discrete variables $\mathtt{\{0, 1\}}$ which determine a linear combination of three 2-dimensional global optimisation benchmark functions (beale, six-hump camel and rosenbrock) and 2 continuous variables $\mathtt{\{2, 3\}}$ which are the input for the given functions. 
    \vspace{-3pt}
    \item \textbf{DTWine} is a hyperparameter optimization problem of decision tree\footnotemark[2] (DT) for classification task on Wine dataset\,\cite{Dua:2019} where the hyperparameters are composed of 2 discrete variables $\mathtt{\{0:splitter, 1:criterion\}}$  and 2 continuous variables $\mathtt{\{2:min\_samples\_split,}$ $\mathtt{3:max\_features\}}$.
    \vspace{-3pt}
    \item \textbf{SVMBoston-4} and \textbf{SVMBoston-5} are hyperparameter optimization problems of nu support vector machine\footnotemark[2] (SVM) for regression task on Boston dataset\,\cite{Dua:2019} where the numbers denote the dimension of the task; the hyperparameters contain 2 discrete variables $\mathtt{\{0:kernel, 1:gamma\}}$ and 2 continuous variables $\mathtt{\{2:C, 3:nu\}}$. One additional hyperparameter for 5-dimensional task is $\mathtt{\{4:tolerance\}}$. 
\end{itemize}
\vspace{-10pt}
\footnotetext[1]{The numbers in the braces are the node indices assigned to each variable for means of concise indicator in the result figures.}
\footnotetext[2]{\url{https://scikit-learn.org/}}

The optimization process is described in Algorithm \ref{alg:prior}. In order to make information of all variables to be propagated within a graph, we ensure the graphs to be \textit{connected}. We compare our approach with the current solid baseline model in the mixed-space optimization literature, CoCaBO\,\cite{ru2020bayesian}. The experimental details are discussed in \autoref{expsetting}.

\autoref{toyproof} demonstrates instances of our approach’s high performance compared to the baseline model. Over all tasks, the application of a proper graph results in not only fast convergence but also an improvement in final performance. To investigate the correlation between the graph’s capability of expressing contributable interactions and how each variable is positioned in the graph structure, we computed Pearson correlation coefficients with the average performance values and PR values for each node. As presented on the right side of the heatmaps in \autoref{toyproof}, we can identify that the importance of certain nodes (inputs) within a graph has a high correlation with the graph's performance, i.e. the more the input interacts with others, the better the optimization performance tends to be, and vice versa. Such distinct relationships appear across all tasks, and such nodes are marked in bold. In other words, it can be inferred that the appropriate graph structure to be used for interaction modeling for optimization contains \textit{hub} nodes. 

A comparison of SVMBoston-4 and SVMBoston-5 explains how the featured relationships change as the task scales by an additional configuration. The $\mathtt{node\:1}$, or $\mathtt{gamma}$, maintains its positive correlation, whereas the $\mathtt{node\:2}$ and $\mathtt{node\:3}$, or $\mathtt{C}$ and $\mathtt{nu}$, preserves a distinct negative correlation. In constrast, the $\mathtt{node\:0}$, or $\mathtt{kernel}$, appears to have the highest positive correlation in the latter task, while it shows irrelevance in the former. Nevertheless, the comparison shows that the general relationship, either positive or negative, stays consistent regardless of additional input, while the superiority in degree is flexible. Thus, the existence of relatively \textit{important} nodes and unimportant nodes is not a coincidence, but a coherent pattern observed among the inputs. In the interest of finding optimal connectivities in unseen tasks, we describe our development of an adapted mechanism of graph structure learning in the next section.

\vspace{-8pt}
\subsection{Proposed Algorithm}
\label{proposed}

We present a new graph structure learning module, referred to as \textit{nested} MAB, which operates two agents of Multi-Armed Bandit\,(MAB). For both MABs, EXP3\,\cite{auer2003} is adopted because it makes relatively fewer assumptions on reward distributions. Combined with the described LSO framework, it establishes our proposed method GEBO, as illustrated in \autoref{gebo}.
\begin{figure}[t]
\begin{center}
\centerline{\includegraphics[width=\columnwidth]{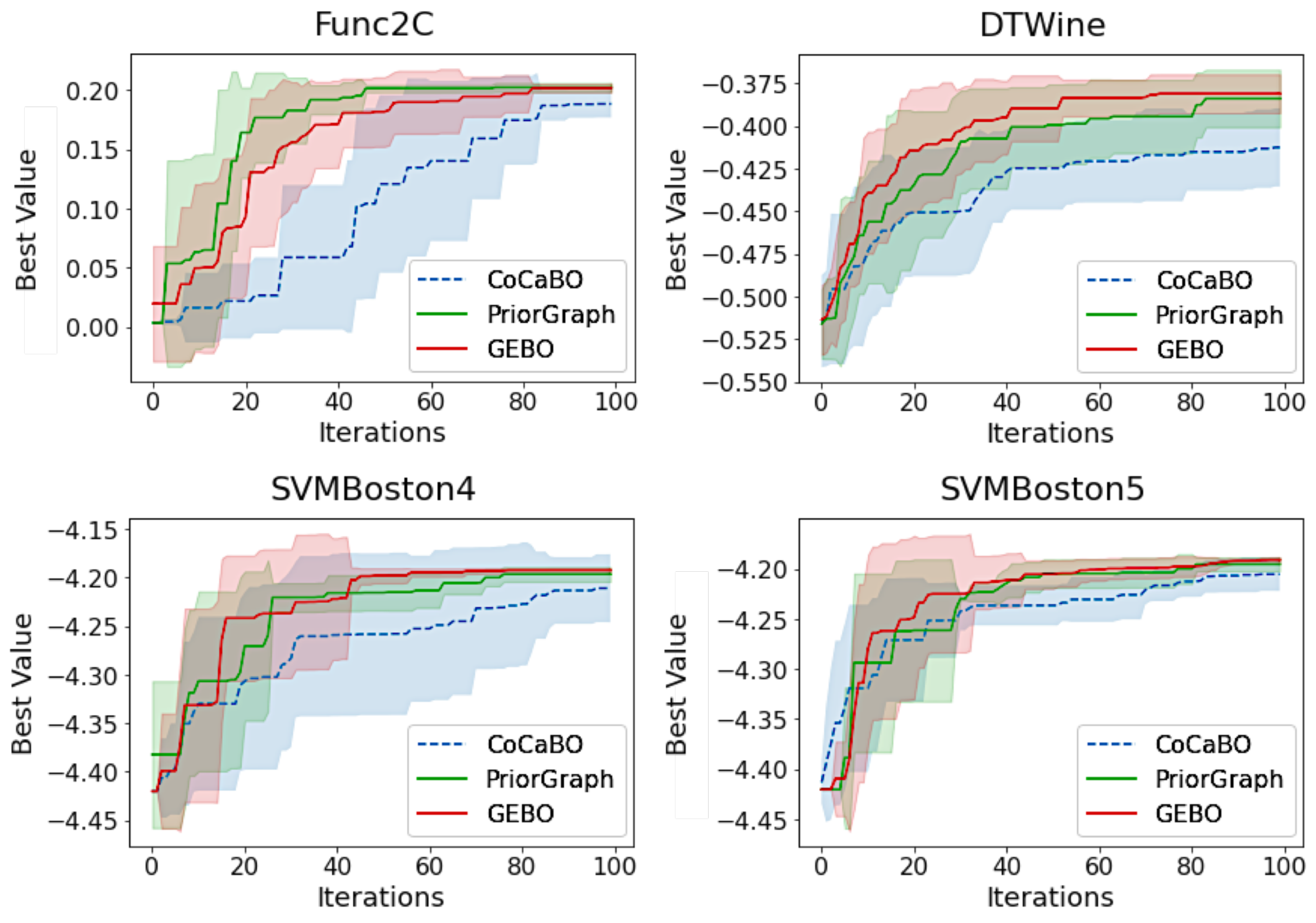}}
\caption{Proposed GEBO approach (red solid line) plotted by mean $\pm$ 1 standard deviation along with high-performing graph example (green solid line) and baseline model CoCaBO (blue dashed line).} 
\label{validate}
\end{center}
\vskip -0.2in
\end{figure}

\begin{algorithm}[tb]
   \caption{GEBO (Proposed Algorithm)}
   \label{alg:proposed}
\begin{algorithmic}
    \STATE {\bfseries Input:} initial observations $\mathcal{X}_{init}$, budget $T$, generative/inverse model $g$/$q$, surrogate model $\mathcal{M}$
    \FOR{$k=1$ {\bfseries to} $K$} 
	    \STATE Sample node $c$ times with replacement $\mathcal{V}_{k} \leftarrow Agent_{\mathcal{N}}$
	    \STATE Generate a graph centered by the sampled nodes $Graph_{k}\{\mathcal{V}_{k}\} \leftarrow$ BA-$biased(\mathcal{V}_{k})$
	\ENDFOR
    \STATE Initialize $\mathcal{G} = [Graph_{1}\{\mathcal{V}_{1}\}, \dots , Graph_{K}\{\mathcal{V}_{K}\}]$ 
    \STATE Initialize $\mathcal{X}_{0} = \mathcal{X}_{init}$ and warm-up train $g$/$q$
    \FOR{$t=1$ {\bfseries to} $T$}
        \STATE Select $Graph_{i} \leftarrow Agent_{\mathcal{G}}$ s.t. $i \in [1, K]$
        \STATE Compute next candidate $\mathbf{x^{t}}$ through LSO and query its evaluation $f(\mathbf{x^{t}})$ (see Algorithm \ref{alg:prior})
        \STATE Update $Agent_{\mathcal{G}}$ with reward $R$ for $i^{th}$ expert and $Agent_{\mathcal{N}}$ with reward $R/|\mathcal{V}_{i}|$ for nodes in $\mathcal{V}_{i}$
        \STATE Retrain generative/inverse model $g$/$q$ with $\mathcal{D}_{t}$
        \IF{Successive failure of $Graph_{j}, \forall j \in [1,K]$}
            \STATE Replace it with new graph generated by updated MABs and reset weight of $j^{th}$arm of $Agent_{\mathcal{G}}$ to 1.
        \ENDIF
    \ENDFOR
    \end{algorithmic}
\end{algorithm}
\setlength{\textfloatsep}{10pt}
To identify \textit{hub} nodes, the first agent, $Agent_{\mathcal{N}}$, is instantiated with inputs (nodes) as its arms, i.e. given $n$-dimensional tasks, it has $n$ number of arms. The agent selects a set of nodes $\mathcal{V}$ that are assumed to be \textit{important} by pulling its arms $c$ times independently. Consequently, the number of nodes is flexibly determined, with one being the minimum and $c$ being the maximum. Then, a random graph is generated, in which the sampled nodes are centered. For the random graph generation, we use the classic random graph generator, Barab\'{a}si-Albert (BA) model\,\cite{RevModPhys.74.47}. 

We introduce a modification to the model\footnote[3]{The original model starts from randomly selected, disconnected set of $m$ nodes and connect newly added nodes each with $m$ edges where the probability of connection to the existing node $v$ is proportional to $v$'s degree.} wherein we start with a \textit{complete} graph of the sampled nodes and add the other nodes sequentially, each with a \textit{single} edge attached to the existing node, where the probability of connection to node $v$ is proportional to $v$'s degree. Thus, our modified model, referred to as BA-\textit{biased}, draws a biased graph on a given prior, i.e. a graph that is highly centered around a predefined set of nodes while the other nodes have relatively sparse connections. By repeating this process $K$ times, we can obtain $K$ different randomly centered graphs. With the graphs initiated as arms of the second agent, $Agent_{\mathcal{G}}$, at the beginning of every iteration, $Agent_{\mathcal{G}}$ selects the graph $i\in[1,K]$ to be encoded for the current iteration. After obtaining a new candidate input $\mathbf{x^{t}}$ by LSO, the evaluation $f(\mathbf{x^{t}})$ is then used to update the reward distribution of the selected graph. Consequently, the reward is also distributed to the reward distribution of the centered nodes of the graph, i.e.
\begin{equation}
\begin{gathered}
    \omega^{\mathcal{G}}_{i}(t+1)=\omega^{\mathcal{G}}_{i}(t)e^{\gamma\hat{f}_{\mathcal{G}}(\mathbf{x^{t}})/K} \\
    \omega^{\mathcal{N}}_{v}(t+1)=\omega^{\mathcal{N}}_{v}(t)e^{\gamma\hat{f}_{\mathcal{N}v}(\mathbf{x^{t}})/(|\mathcal{V}_{i}| \cdot K)},\:\forall{v}\in\mathcal{V}_{i} \\
    \textrm{s.t.}\:\hat{f}_{\mathcal{G}}(\mathbf{x^{t}})=f(\mathbf{x^{t}})/p^{\mathcal{G}}_{it}(t), \\ \hat{f}_{\mathcal{N}v}(\mathbf{x^{t}})=\hat{f}_{\mathcal{G}}(\mathbf{x^{t}})/\sum\limits_{j=1}^{K}\mathbbm{1}(v\in\mathcal{V}_{j}) \cdot p^{\mathcal{G}}_{jt}(t)
\end{gathered}
\end{equation}

where $\hat{f}_{\mathcal{G}}(\mathbf{x^{t}})$ and $\hat{f}_{\mathcal{N}v}(\mathbf{x^{t}})$ are estimated rewards for $Agent_{\mathcal{G}}$ and node $v$ of $Agent_{\mathcal{N}}$ respectively, and $p^{\mathcal{G}}_{it}(t)$ is the probability distribution of graph $i$ at iteration $t$. Note that the rewards for centered nodes $v\in\mathcal{V}_i$ are not only divided by its number $|\mathcal{V}_i|$, but further revised by the probability of performing as a centered node in the current iteration. Thus, it functions as a \textit{nested} MAB, where the reward for a graph is \textit{nested} to the nodes. When a certain graph successively fails to make improvement, the graph is replaced by the updated MABs and the corresponding arm's weight is reset to 1, i.e. $\omega^{\mathcal{G}}_{j}(t+1)=1$ for replaced graph $j$. For every replacement, we normalize the weights by making the sum to be $K$, i.e. the new graph would have a probability of being selected of $1/K$. In the sense that we are assigning reward both to action\,(graph) and meta-action\,(node), our algorithm has analogy with EXP4\,\cite{auer2003}. However, instead of adopting probability distribution over actions (graphs), we preserve the discrete graph search space and sample a set of actions with the use of random graph generator.

\begin{figure}[t]
\begin{center}
\centerline{\includegraphics[width=\columnwidth]{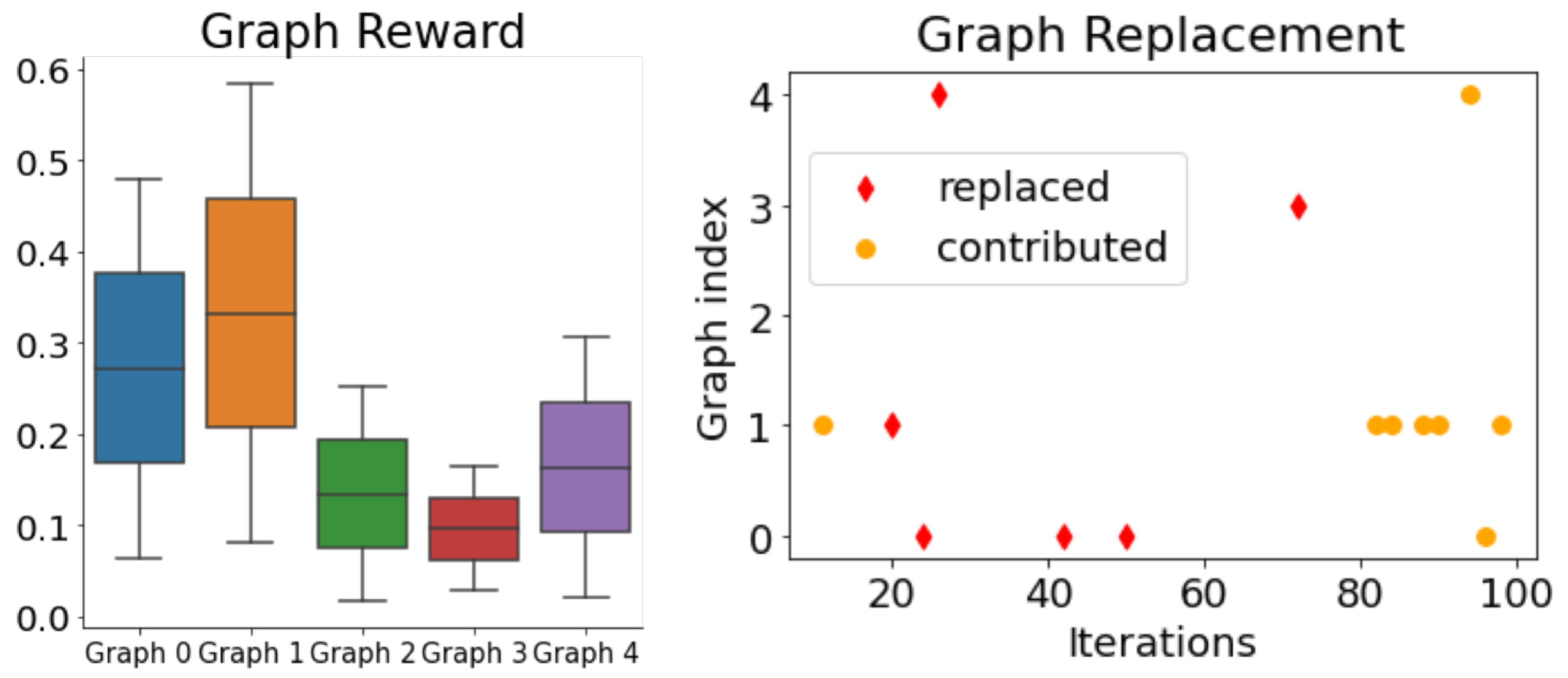}}
\caption{(Left) Average of the graph reward at the end of optimization; (Right) Indices of graphs that are replaced (red diamond) or have contributed to the optimization performance (orange dot).}
\label{reward}
\end{center}
\vskip -0.2in
\end{figure}

We evaluate our graph learning mechanism on the four previous tasks to validate the search efficiency. As \autoref{validate} shows, \textit{nested} MAB successfully converges to the performance level of the high-performing graph over the iterations.  \autoref{reward} shows the effectiveness of our algorithm on the task of SVMBoston-5. The average graph reward and track of graph indices that are either replaced or contributed not only show that the replacing mechanism effectively regenerates contributing graphs but also that our algorithm utilizes different graphs, i.e. perspectives, for optimization, which explains why its performance often exceeds that of the prior graph.  

\begin{figure}[t!]
\vskip 0.1in
\begin{center}
\centerline{\includegraphics[width=\columnwidth]{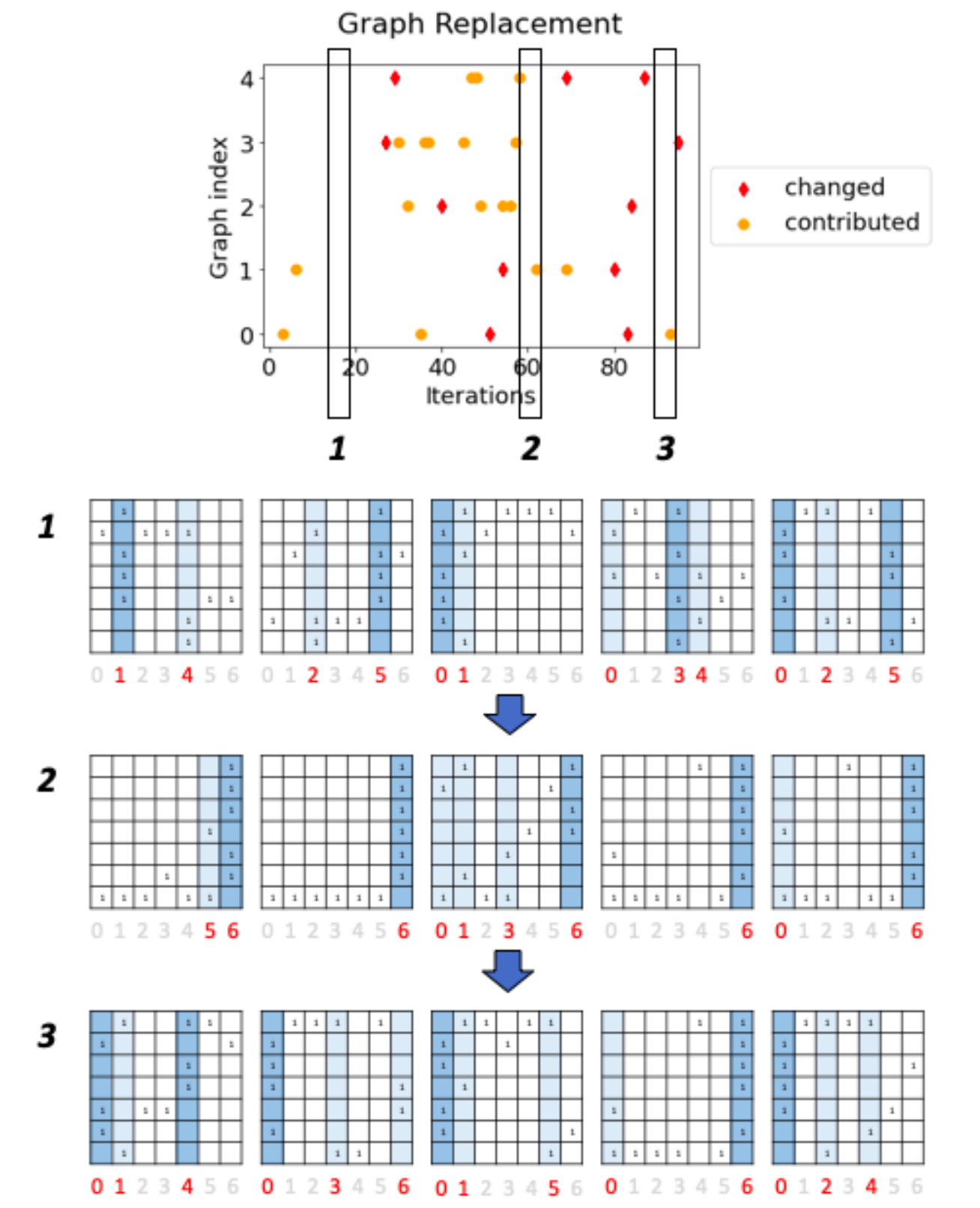}}
\caption{Track of graph indices and visualization of the graphs in different phases for the speed reducer design task. The centered nodes are colored in blue, and the centrality is discriminated by the darkness. In the first phase, the graphs are generated with a diverse set of centered nodes. After replacements, the generated graphs tend to focus on the $\mathtt{node\:6}$. In the last phase, the centered nodes converge to $\mathtt{node\:0}$, $\mathtt{4}$, and $\mathtt{6}$.}
\label{graphchange}
\end{center}
\vskip -0.2in
\end{figure}

\begin{figure*}[ht]
\vskip 0.2in
\begin{center}
\centerline{\includegraphics[width=\textwidth]{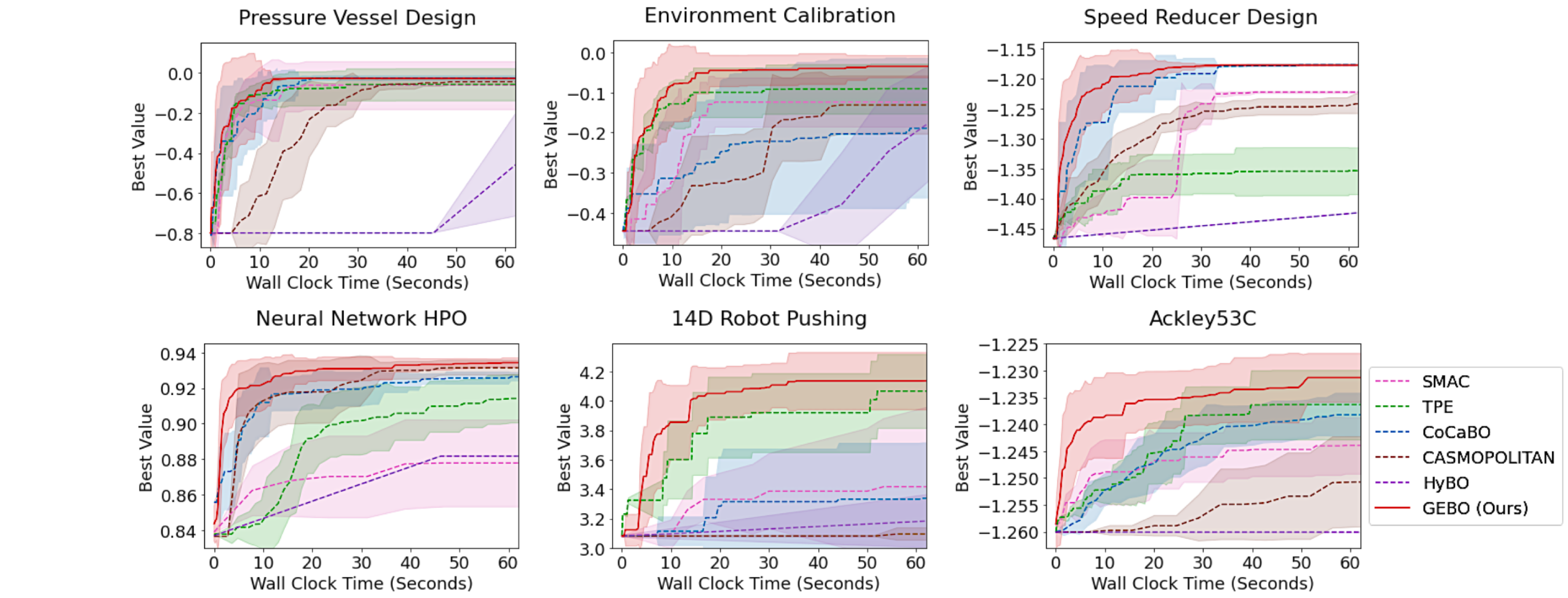}}
\caption{Performance reported by mean $\pm$ standard deviation over 10 random repetitions compared in terms of computation time. Within the limited time of 60 seconds, GEBO outperforms the baseline methods with fast convergence for all tasks.  We include additional performance comparison over iterations in \autoref{appendixexp}.}
\label{mainexp}
\end{center}
\vskip -0.2in
\vspace{-5pt}
\end{figure*}

\vspace{-5pt}
\section{Experiment}
\vspace{-5pt}
We evaluate GEBO on synthetic and real-world tasks which involve more challenges that align with the actual problem scenarios. We consider the following problems:
\vspace{-10pt}
\begin{itemize}
    \item \textbf{Pressure vessel design} is a problem of designing cylindrical pressure vessel to minimize the required cost\,\cite{kannan1994, TANABE2020106078} and the task involves 2 discrete (thickness of shell and head of pressure vessel) and 2 continuous (inner radius and length of cylindrical section) variables. Both of the discrete variables have 100 possible values each.
    \vspace{-5pt}
    \item \textbf{Environment calibration} consists of 1 discrete and 3 continuous variables where the discrete variable has 285 dimensions. The configurations determine expensive scientific environmental models for chemical accident\,\cite{bliznyuk2008env, pmlr-v97-astudillo19a}.
    \vspace{-18pt}
    \item \textbf{Speed reducer design} is a mechanical design problem for speed reducer where the objective is minimizing the weight\,\cite{GOLINSKI1973419, cagnina2008}. It contains 1 discrete variable (number of teeth on pinion) with 12 dimensions and 6 continuous (face width, teeth module, lengths of shafts between bearings, and diameters of the shafts) variables. 
    \vspace{-5pt}
    \item \textbf{Neural network hyperparameter optimization (HPO)} requires tuning of MLP classifier model on automl benchmark\,\cite{gijsbers2019open}. We tested on segment dataset and the configuration is composed of 4 discrete (hidden
    layer size, activation type, and type of learning rate) and 3 continuous (learning rate initialization, momentum parameter, and regularization coefficient) inputs. The discrete variables have 14, 4, 9, and 3 dimensions respectively.
    \vspace{-5pt}
    \item \textbf{14D Robot pushing} is a control parameter optimization task of robot pushing toward the target location\,\cite{wang2018batched, deshwal2021bayesian}. The search space contains 10 discrete variables (4 of 11-dimensional, 4 of 21-dimensional, and 2 of 29-dimensional; determines the location and number of simulation steps) and 4 continuous variables (determines rotation).
    \vspace{-5pt}
    \item \textbf{Ackley53C} is a synthetic problem of Ackley function\,\cite{bliek2021, wan2021think} consisting 50 discrete variables with 2 dimensions and 3 continuous variables.
\end{itemize}
\vspace{-10pt}
We compare our algorithm with 5 strong baseline models: SMAC\,\cite{hutter2011sequential}, TPE\,\cite{bergstra2011algorithms}, CoCaBO\,\cite{ru2020bayesian}, CASMOPOLITAN\,\cite{wan2021think}, and HyBO\,\cite{deshwal2021bayesian}. For all experiments, a fixed hyperparameter setting of $K=5$ and $c=3$ is used and GP-BO is conducted with Mat$\acute{e}$rn kernel for the LSO. We run the experiments on AMD Ryzen 9 3900X 12-Core machine with 1 RTX 2080Ti GPU card. Further details of the experiment setting are discussed in \autoref{expsetting}.

\subsection{Results}
\autoref{mainexp} shows the experiment results reported with the $x$-axis of time required for computing the next suggestion by wall-clock time in seconds. For all tasks, GEBO outperforms the existing methods, showing fast convergence and high performance within the limited time budget of 60 seconds. CoCaBO converges relatively slowly when high-dimensional discrete variables are involved, often underperforming, due to the MAB's need to explore large discrete subspaces. CASMOPOLITAN consumes a relatively large amount of computation time for its interleaving search mechanism between separated subspaces, which in consequence resulted in low performance with respect to time. On the other hand, HyBO barely completes its first evaluation within the constrained time frame, which is more clearly shown in high-dimensional problems. SMAC and TPE generally underperforms in all tasks with high variance, which can be explained by inaccuracy from its randomness and exclusion of interaction learning, respectively. Above all, GEBO consistently shows outstanding performance, effectively searching for crucial nodes (inputs) and generating better graphs over the iterations that contributed to target optimization as shown in \autoref{graphchange}.


\begin{figure}[h]
\begin{center}
\centerline{\includegraphics[width=\columnwidth]{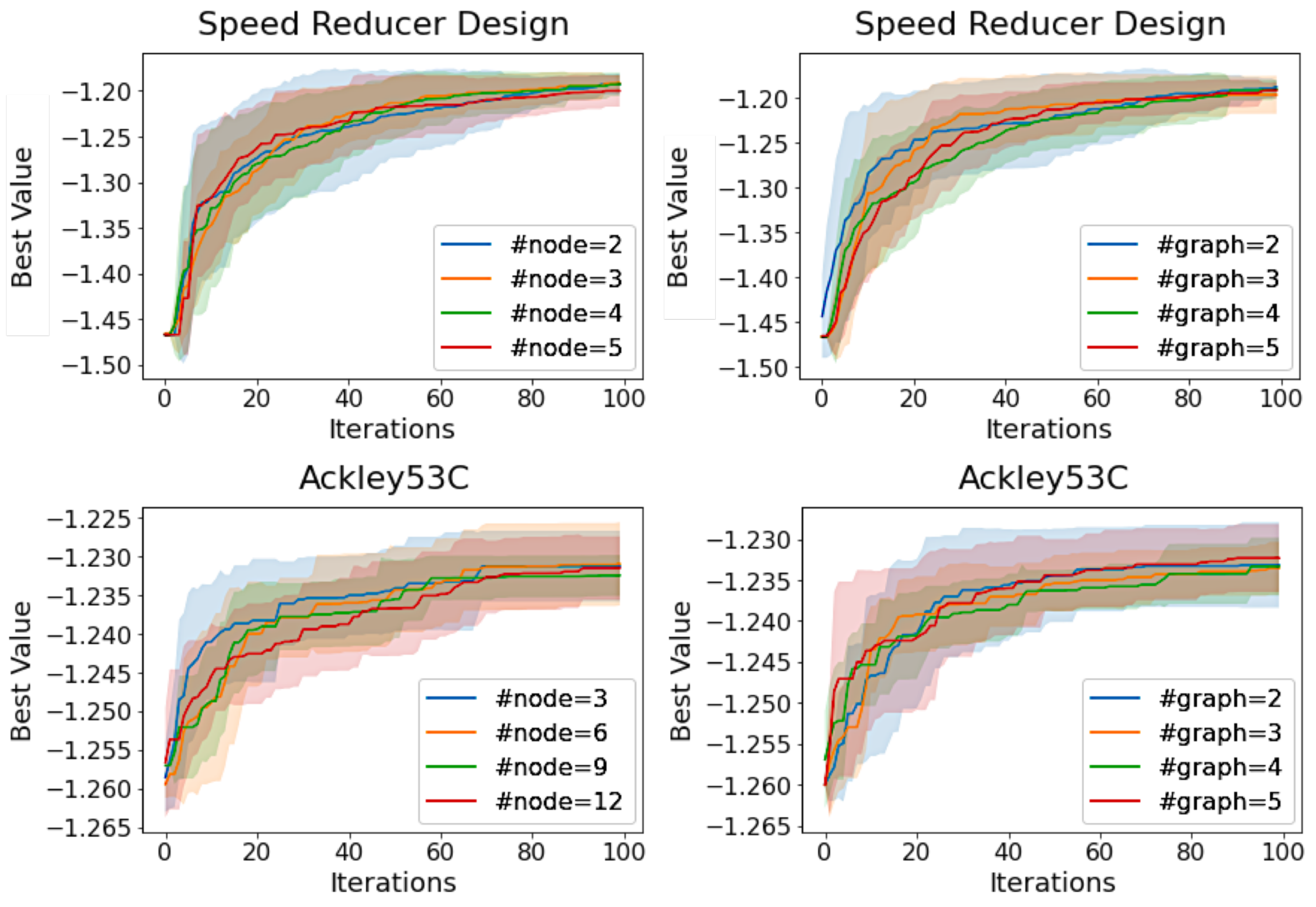}}
\caption{Ablation studies of GEBO on two hyperparameters: number of nodes to be centered in a graph ($c$) and number of graphs to generate ($K$). For the former comparison, we fixed $K=5$ and for the latter, fixed value of $c=3$ is used.}
\label{ablation}
\end{center}
\vskip -0.2in
\end{figure}

\vspace{-10pt}

\subsection{Ablation Studies}
\vspace{-5pt}

We conduct ablation studies on hyperparameter settings to demonstrate the robustness of GEBO, as shown in \autoref{ablation}. The different values induce a small gap in the initial stage of the optimization process, but in the end, they all converge into an equivalently high level of performance. We can infer that the flexibility of the number of centered nodes derived from sampling with replacement alleviates the possible variance caused by different choices of values. In addition, regardless of the number of graphs in hand, the graph replacement mechanism efficiently creates more suitable graph structures over iterations.

\vspace{-10pt}
\section{Conclusion}
We propose a novel graph-based approach, GEBO, for mixed-space optimization which forms a joint framework of graph structure learning and LSO. The key idea is to mold the data into a \textit{graph} to model the complex interactions between mixed inputs so that with VGAE, it allows not only the natural integration of relational information but also utilization of GP-BO in the learned latent space. GEBO shows effective performance over diverse tasks, outperforming the state-of-the-art methods in computational efficiency. Future works are expected on probabilistic modeling of graph search spaces for differentiable graph sampling.

\newpage
\bibliography{main}

\begin{thebibliography}{54}
\providecommand{\natexlab}[1]{#1}
\providecommand{\url}[1]{\texttt{#1}}
\expandafter\ifx\csname urlstyle\endcsname\relax
  \providecommand{\doi}[1]{doi: #1}\else
  \providecommand{\doi}{doi: \begingroup \urlstyle{rm}\Url}\fi

\bibitem[Abbassi \& Mirrokni(2007)Abbassi and Mirrokni]{Zeinab}
Abbassi, Z. and Mirrokni, V.~S.
\newblock A recommender system based on local random walks and spectral
  methods.
\newblock New York, NY, USA, 2007. Association for Computing Machinery.
\newblock ISBN 9781595938480.
\newblock \doi{10.1145/1348549.1348561}.
\newblock URL \url{https://doi.org/10.1145/1348549.1348561}.

\bibitem[Albert \& Barab\'asi(2002)Albert and Barab\'asi]{RevModPhys.74.47}
Albert, R. and Barab\'asi, A.-L.
\newblock Statistical mechanics of complex networks.
\newblock \emph{Rev. Mod. Phys.}, 74:\penalty0 47--97, Jan 2002.
\newblock \doi{10.1103/RevModPhys.74.47}.
\newblock URL \url{https://link.aps.org/doi/10.1103/RevModPhys.74.47}.

\bibitem[Astudillo \& Frazier(2019)Astudillo and
  Frazier]{pmlr-v97-astudillo19a}
Astudillo, R. and Frazier, P.
\newblock {B}ayesian optimization of composite functions.
\newblock In Chaudhuri, K. and Salakhutdinov, R. (eds.), \emph{Proceedings of
  the 36th International Conference on Machine Learning}, volume~97 of
  \emph{Proceedings of Machine Learning Research}, pp.\  354--363. PMLR, 09--15
  Jun 2019.
\newblock URL \url{https://proceedings.mlr.press/v97/astudillo19a.html}.

\bibitem[Audet \& Hare(2017)Audet and Hare]{audet2017derivative}
Audet, C. and Hare, W.
\newblock Derivative-free and blackbox optimization.
\newblock 2017.

\bibitem[Auer et~al.(2003)Auer, Cesa-Bianchi, Freund, and Schapire]{auer2003}
Auer, P., Cesa-Bianchi, N., Freund, Y., and Schapire, R.
\newblock The nonstochastic multiarmed bandit problem.
\newblock \emph{SIAM Journal on Computing}, 32:\penalty0 48--77, 01 2003.

\bibitem[Bajaj et~al.(2021)Bajaj, Arora, and Hasan]{Bajaj2021}
Bajaj, I., Arora, A., and Hasan, M. M.~F.
\newblock \emph{Black-Box Optimization: Methods and Applications}, pp.\
  35--65.
\newblock Springer International Publishing, Cham, 2021.
\newblock ISBN 978-3-030-66515-9.
\newblock \doi{10.1007/978-3-030-66515-9_2}.
\newblock URL \url{https://doi.org/10.1007/978-3-030-66515-9_2}.

\bibitem[Bergstra et~al.(2011)Bergstra, Bardenet, Bengio, and
  K{\'e}gl]{bergstra2011algorithms}
Bergstra, J., Bardenet, R., Bengio, Y., and K{\'e}gl, B.
\newblock Algorithms for hyper-parameter optimization.
\newblock \emph{Advances in neural information processing systems}, 24, 2011.

\bibitem[Bliek et~al.(2021)Bliek, Guijt, Verwer, and de~Weerdt]{bliek2021}
Bliek, L., Guijt, A., Verwer, S., and de~Weerdt, M.
\newblock Black-box mixed-variable optimisation using a surrogate model that
  satisfies integer constraints.
\newblock \emph{Proceedings of the Genetic and Evolutionary Computation
  Conference Companion}, Jul 2021.
\newblock \doi{10.1145/3449726.3463136}.
\newblock URL \url{http://dx.doi.org/10.1145/3449726.3463136}.

\bibitem[Bliznyuk et~al.(2008)Bliznyuk, Ruppert, Shoemaker, Regis, Wild, and
  Mugunthan]{bliznyuk2008env}
Bliznyuk, N., Ruppert, D., Shoemaker, C., Regis, R., Wild, S., and Mugunthan,
  P.
\newblock Bayesian calibration and uncertainty analysis for computationally
  expensive models using optimization and radial basis function approximation.
\newblock \emph{Journal of Computational and Graphical Statistics - J COMPUT
  GRAPH STAT}, 17:\penalty0 270--294, 06 2008.
\newblock \doi{10.1198/106186008X320681}.

\bibitem[Breiman(2001)]{breiman2001random}
Breiman, L.
\newblock Random forests.
\newblock \emph{Machine learning}, 45\penalty0 (1):\penalty0 5--32, 2001.

\bibitem[Cagnina et~al.(2008)Cagnina, Esquivel, and Coello]{cagnina2008}
Cagnina, L., Esquivel, S., and Coello, C.
\newblock Solving engineering optimization problems with the simple constrained
  particle swarm optimizer.
\newblock \emph{Informatica (Slovenia)}, 32:\penalty0 319--326, 01 2008.

\bibitem[Deshwal et~al.(2021)Deshwal, Belakaria, and
  Doppa]{deshwal2021bayesian}
Deshwal, A., Belakaria, S., and Doppa, J.~R.
\newblock Bayesian optimization over hybrid spaces.
\newblock In Meila, M. and Zhang, T. (eds.), \emph{Proceedings of the 38th
  International Conference on Machine Learning, {ICML} 2021, 18-24 July 2021,
  Virtual Event}, volume 139 of \emph{Proceedings of Machine Learning
  Research}, pp.\  2632--2643. {PMLR}, 2021.
\newblock URL \url{http://proceedings.mlr.press/v139/deshwal21a.html}.

\bibitem[Dua \& Graff(2017)Dua and Graff]{Dua:2019}
Dua, D. and Graff, C.
\newblock {UCI} machine learning repository, 2017.
\newblock URL \url{http://archive.ics.uci.edu/ml}.

\bibitem[Duvenaud et~al.(2011)Duvenaud, Nickisch, and
  Rasmussen]{duvenaud2011additive}
Duvenaud, D., Nickisch, H., and Rasmussen, C.~E.
\newblock Additive gaussian processes, 2011.

\bibitem[Elsken et~al.(2019)Elsken, Metzen, and Hutter]{elsken2019neural}
Elsken, T., Metzen, J.~H., and Hutter, F.
\newblock Neural architecture search: A survey, 2019.

\bibitem[Elton et~al.(2019)Elton, Boukouvalas, Fuge, and
  Chung]{elton2019molecular}
Elton, D.~C., Boukouvalas, Z., Fuge, M.~D., and Chung, P.~W.
\newblock Deep learning for molecular design—a review of the state of the
  art.
\newblock \emph{Mol. Syst. Des. Eng.}, 4:\penalty0 828--849, 2019.
\newblock \doi{10.1039/C9ME00039A}.
\newblock URL \url{http://dx.doi.org/10.1039/C9ME00039A}.

\bibitem[Eriksson et~al.(2019)Eriksson, Pearce, Gardner, Turner, and
  Poloczek]{eriksson2020scalable}
Eriksson, D., Pearce, M., Gardner, J., Turner, R.~D., and Poloczek, M.
\newblock Scalable global optimization via local bayesian optimization.
\newblock In Wallach, H., Larochelle, H., Beygelzimer, A., d\textquotesingle
  Alch\'{e}-Buc, F., Fox, E., and Garnett, R. (eds.), \emph{Advances in Neural
  Information Processing Systems}, volume~32. Curran Associates, Inc., 2019.
\newblock URL
  \url{https://proceedings.neurips.cc/paper/2019/file/6c990b7aca7bc7058f5e98ea909e924b-Paper.pdf}.

\bibitem[Frazier(2018)]{frazier2018tutorial}
Frazier, P.~I.
\newblock A tutorial on bayesian optimization, 2018.

\bibitem[Gijsbers et~al.(2019)Gijsbers, LeDell, Thomas, Poirier, Bischl, and
  Vanschoren]{gijsbers2019open}
Gijsbers, P., LeDell, E., Thomas, J., Poirier, S., Bischl, B., and Vanschoren,
  J.
\newblock An open source automl benchmark, 2019.

\bibitem[Golinski(1973)]{GOLINSKI1973419}
Golinski, J.
\newblock An adaptive optimization system applied to machine synthesis.
\newblock \emph{Mechanism and Machine Theory}, 8\penalty0 (4):\penalty0
  419--436, 1973.
\newblock ISSN 0094-114X.
\newblock \doi{https://doi.org/10.1016/0094-114X(73)90018-9}.
\newblock URL
  \url{https://www.sciencedirect.com/science/article/pii/0094114X73900189}.

\bibitem[Grolmusz(2015)]{grolmusz2015}
Grolmusz, V.
\newblock A note on the pagerank of undirected graphs.
\newblock \emph{Information Processing Letters}, 115\penalty0 (6-8):\penalty0
  633–634, Jun 2015.
\newblock ISSN 0020-0190.
\newblock \doi{10.1016/j.ipl.2015.02.015}.
\newblock URL \url{http://dx.doi.org/10.1016/j.ipl.2015.02.015}.

\bibitem[Grosnit et~al.(2021)Grosnit, Tutunov, Maraval, Griffiths,
  Cowen-Rivers, Yang, Zhu, Lyu, Chen, Wang, et~al.]{grosnit2021high}
Grosnit, A., Tutunov, R., Maraval, A.~M., Griffiths, R.-R., Cowen-Rivers,
  A.~I., Yang, L., Zhu, L., Lyu, W., Chen, Z., Wang, J., et~al.
\newblock High-dimensional bayesian optimisation with variational autoencoders
  and deep metric learning.
\newblock \emph{arXiv preprint arXiv:2106.03609}, 2021.

\bibitem[Hutter et~al.(2011)Hutter, Hoos, and
  Leyton-Brown]{hutter2011sequential}
Hutter, F., Hoos, H.~H., and Leyton-Brown, K.
\newblock Sequential model-based optimization for general algorithm
  configuration.
\newblock In \emph{International conference on learning and intelligent
  optimization}, pp.\  507--523. Springer, 2011.

\bibitem[Iván \& Grolmusz(2011)Iván and Grolmusz]{Gabor}
Iván, G. and Grolmusz, V.
\newblock When the web meets the cell: Using personalized pagerank for
  analyzing protein interaction networks.
\newblock \emph{Bioinformatics (Oxford, England)}, 27:\penalty0 405--7, 02
  2011.
\newblock \doi{10.1093/bioinformatics/btq680}.

\bibitem[Jones et~al.(1998)Jones, Schonlau, and Welch]{jones1998expensivebbo}
Jones, D., Schonlau, M., and Welch, W.
\newblock Efficient global optimization of expensive black-box functions.
\newblock \emph{Journal of Global Optimization}, 13:\penalty0 455--492, 12
  1998.
\newblock \doi{10.1023/A:1008306431147}.

\bibitem[Kannan \& Kramer(1994)Kannan and Kramer]{kannan1994}
Kannan, B.~K. and Kramer, S.~N.
\newblock {An Augmented Lagrange Multiplier Based Method for Mixed Integer
  Discrete Continuous Optimization and Its Applications to Mechanical Design}.
\newblock \emph{Journal of Mechanical Design}, 116\penalty0 (2):\penalty0
  405--411, 06 1994.
\newblock ISSN 1050-0472.
\newblock \doi{10.1115/1.2919393}.
\newblock URL \url{https://doi.org/10.1115/1.2919393}.

\bibitem[Kim et~al.(2019)Kim, Seo, Laptev, Cho, and Kwak]{Kim2019DeepML}
Kim, S., Seo, M., Laptev, I., Cho, M., and Kwak, S.
\newblock Deep metric learning beyond binary supervision.
\newblock \emph{2019 IEEE/CVF Conference on Computer Vision and Pattern
  Recognition (CVPR)}, pp.\  2283--2292, 2019.

\bibitem[Kipf \& Welling(2016)Kipf and Welling]{kipf2016variational}
Kipf, T.~N. and Welling, M.
\newblock Variational graph auto-encoders, 2016.

\bibitem[Mei et~al.(2016)Mei, Omidvar, Li, and Yao]{mei2016competitive}
Mei, Y., Omidvar, M.~N., Li, X., and Yao, X.
\newblock A competitive divide-and-conquer algorithm for unconstrained
  large-scale black-box optimization.
\newblock \emph{ACM Transactions on Mathematical Software (TOMS)}, 42\penalty0
  (2):\penalty0 1--24, 2016.

\bibitem[Notin et~al.(2021)Notin, Hern{\'a}ndez-Lobato, and
  Gal]{notin2021improving}
Notin, P., Hern{\'a}ndez-Lobato, J.~M., and Gal, Y.
\newblock Improving black-box optimization in vae latent space using decoder
  uncertainty.
\newblock \emph{Advances in Neural Information Processing Systems}, 34, 2021.

\bibitem[Page et~al.(1999)Page, Brin, Motwani, and Winograd]{page1999}
Page, L., Brin, S., Motwani, R., and Winograd, T.
\newblock The pagerank citation ranking: Bringing order to the web.
\newblock Technical Report 1999-66, Stanford InfoLab, November 1999.
\newblock URL \url{http://ilpubs.stanford.edu:8090/422/}.
\newblock Previous number = SIDL-WP-1999-0120.

\bibitem[Perra \& Fortunato(2008)Perra and Fortunato]{Nicola}
Perra, N. and Fortunato, S.
\newblock Spectral centrality measures in complex networks.
\newblock \emph{Physical Review E}, 78\penalty0 (3), Sep 2008.
\newblock ISSN 1550-2376.
\newblock \doi{10.1103/physreve.78.036107}.
\newblock URL \url{http://dx.doi.org/10.1103/PhysRevE.78.036107}.

\bibitem[Rasmussen(2004)]{Rasmussen2004}
Rasmussen, C.~E.
\newblock \emph{Gaussian Processes in Machine Learning}, pp.\  63--71.
\newblock Springer Berlin Heidelberg, Berlin, Heidelberg, 2004.
\newblock ISBN 978-3-540-28650-9.
\newblock \doi{10.1007/978-3-540-28650-9_4}.
\newblock URL \url{https://doi.org/10.1007/978-3-540-28650-9_4}.

\bibitem[Ren et~al.(2021)Ren, Xiao, Chang, Huang, Li, Chen, and
  Wang]{ren2021comprehensive}
Ren, P., Xiao, Y., Chang, X., Huang, P.-Y., Li, Z., Chen, X., and Wang, X.
\newblock A comprehensive survey of neural architecture search: Challenges and
  solutions, 2021.

\bibitem[Ru et~al.(2020)Ru, Alvi, Nguyen, Osborne, and Roberts]{ru2020bayesian}
Ru, B.~X., Alvi, A.~S., Nguyen, V., Osborne, M.~A., and Roberts, S.~J.
\newblock Bayesian optimisation over multiple continuous and categorical
  inputs.
\newblock In \emph{Proceedings of the 37th International Conference on Machine
  Learning, {ICML} 2020, 13-18 July 2020, Virtual Event}, volume 119 of
  \emph{Proceedings of Machine Learning Research}, pp.\  8276--8285. {PMLR},
  2020.
\newblock URL \url{http://proceedings.mlr.press/v119/ru20a.html}.

\bibitem[Sanchez \& Aspuru-Guzik(2018)Sanchez and
  Aspuru-Guzik]{sanchez2018molecular}
Sanchez, B. and Aspuru-Guzik, A.
\newblock Inverse molecular design using machine learning: Generative models
  for matter engineering.
\newblock \emph{Science}, 361:\penalty0 360--365, 07 2018.
\newblock \doi{10.1126/science.aat2663}.

\bibitem[Scarselli et~al.(2009)Scarselli, Gori, Tsoi, Hagenbuchner, and
  Monfardini]{scarselli2009ieee}
Scarselli, F., Gori, M., Tsoi, A.~C., Hagenbuchner, M., and Monfardini, G.
\newblock The graph neural network model.
\newblock \emph{IEEE Transactions on Neural Networks}, 20\penalty0
  (1):\penalty0 61--80, 2009.
\newblock \doi{10.1109/TNN.2008.2005605}.

\bibitem[Shahriari et~al.(2016)Shahriari, Swersky, Wang, Adams, and
  de~Freitas]{shahriari2016}
Shahriari, B., Swersky, K., Wang, Z., Adams, R.~P., and de~Freitas, N.
\newblock Taking the human out of the loop: A review of bayesian optimization.
\newblock \emph{Proceedings of the IEEE}, 104\penalty0 (1):\penalty0 148--175,
  2016.
\newblock \doi{10.1109/JPROC.2015.2494218}.

\bibitem[Shan \& Wang(2010)Shan and Wang]{shan2010survey}
Shan, S. and Wang, G.~G.
\newblock Survey of modeling and optimization strategies to solve
  high-dimensional design problems with computationally-expensive black-box
  functions.
\newblock \emph{Structural and multidisciplinary optimization}, 41\penalty0
  (2):\penalty0 219--241, 2010.

\bibitem[Shi et~al.(2020)Shi, Pi, Xu, Li, Kwok, and Zhang]{bridge2020}
Shi, H., Pi, R., Xu, H., Li, Z., Kwok, J., and Zhang, T.
\newblock Bridging the gap between sample-based and one-shot neural
  architecture search with bonas.
\newblock In Larochelle, H., Ranzato, M., Hadsell, R., Balcan, M.~F., and Lin,
  H. (eds.), \emph{Advances in Neural Information Processing Systems},
  volume~33, pp.\  1808--1819. Curran Associates, Inc., 2020.
\newblock URL
  \url{https://proceedings.neurips.cc/paper/2020/file/13d4635deccc230c944e4ff6e03404b5-Paper.pdf}.

\bibitem[Siivola et~al.(2021)Siivola, Paleyes, Gonz{\'a}lez, and
  Vehtari]{siivola2021good}
Siivola, E., Paleyes, A., Gonz{\'a}lez, J., and Vehtari, A.
\newblock Good practices for bayesian optimization of high dimensional
  structured spaces.
\newblock \emph{Applied AI Letters}, 2\penalty0 (2):\penalty0 e24, 2021.

\bibitem[Snoek et~al.(2012)Snoek, Larochelle, and Adams]{snoek2012practical}
Snoek, J., Larochelle, H., and Adams, R.~P.
\newblock Practical bayesian optimization of machine learning algorithms, 2012.

\bibitem[Tanabe \& Ishibuchi(2020)Tanabe and Ishibuchi]{TANABE2020106078}
Tanabe, R. and Ishibuchi, H.
\newblock An easy-to-use real-world multi-objective optimization problem suite.
\newblock \emph{Applied Soft Computing}, 89:\penalty0 106078, 2020.
\newblock ISSN 1568-4946.
\newblock \doi{https://doi.org/10.1016/j.asoc.2020.106078}.
\newblock URL
  \url{https://www.sciencedirect.com/science/article/pii/S1568494620300181}.

\bibitem[Tripp et~al.(2020)Tripp, Daxberger, and
  Hernández-Lobato]{tripp2020sampleefficient}
Tripp, A., Daxberger, E., and Hernández-Lobato, J.~M.
\newblock Sample-efficient optimization in the latent space of deep generative
  models via weighted retraining, 2020.

\bibitem[Turner et~al.(2021)Turner, Eriksson, McCourt, Kiili, Laaksonen, Xu,
  and Guyon]{turner2021bayesian}
Turner, R., Eriksson, D., McCourt, M., Kiili, J., Laaksonen, E., Xu, Z., and
  Guyon, I.
\newblock Bayesian optimization is superior to random search for machine
  learning hyperparameter tuning: Analysis of the black-box optimization
  challenge 2020, 2021.

\bibitem[Verma \& Chakraborty(2021)Verma and
  Chakraborty]{verma2021uncertaintyaware}
Verma, E. and Chakraborty, S.
\newblock Uncertainty-aware labelled augmentations for high dimensional latent
  space bayesian optimization.
\newblock In \emph{NeurIPS 2021 Workshop on Deep Generative Models and
  Downstream Applications}, 2021.
\newblock URL \url{https://openreview.net/forum?id=C7pY5Wjwk0d}.

\bibitem[Wan et~al.(2021)Wan, Nguyen, Ha, Ru, Lu, and Osborne]{wan2021think}
Wan, X., Nguyen, V., Ha, H., Ru, B.~X., Lu, C., and Osborne, M.~A.
\newblock Think global and act local: Bayesian optimisation over
  high-dimensional categorical and mixed search spaces.
\newblock In Meila, M. and Zhang, T. (eds.), \emph{Proceedings of the 38th
  International Conference on Machine Learning, {ICML} 2021, 18-24 July 2021,
  Virtual Event}, volume 139 of \emph{Proceedings of Machine Learning
  Research}, pp.\  10663--10674. {PMLR}, 2021.
\newblock URL \url{http://proceedings.mlr.press/v139/wan21b.html}.

\bibitem[Wang et~al.(2007)Wang, Liu, and Wang]{Jinghua}
Wang, J., Liu, J., and Wang, C.
\newblock Keyword extraction based on pagerank.
\newblock In Zhou, Z.-H., Li, H., and Yang, Q. (eds.), \emph{Advances in
  Knowledge Discovery and Data Mining}, pp.\  857--864, Berlin, Heidelberg,
  2007. Springer Berlin Heidelberg.
\newblock ISBN 978-3-540-71701-0.

\bibitem[Wang et~al.(2004)Wang, Shan, and Wang]{wang2004mode}
Wang, L., Shan, S., and Wang, G.~G.
\newblock Mode-pursuing sampling method for global optimization on expensive
  black-box functions.
\newblock \emph{Engineering Optimization}, 36\penalty0 (4):\penalty0 419--438,
  2004.

\bibitem[Wang et~al.(2018)Wang, Gehring, Kohli, and Jegelka]{wang2018batched}
Wang, Z., Gehring, C., Kohli, P., and Jegelka, S.
\newblock Batched large-scale bayesian optimization in high-dimensional spaces,
  2018.

\bibitem[Wu et~al.(2021)Wu, Pan, Chen, Long, Zhang, and Yu]{wu2021graphsurvey}
Wu, Z., Pan, S., Chen, F., Long, G., Zhang, C., and Yu, P.~S.
\newblock A comprehensive survey on graph neural networks.
\newblock \emph{IEEE Transactions on Neural Networks and Learning Systems},
  32\penalty0 (1):\penalty0 4--24, 2021.
\newblock \doi{10.1109/TNNLS.2020.2978386}.

\bibitem[Xie et~al.(2019)Xie, Kirillov, Girshick, and He]{xie2019exploring}
Xie, S., Kirillov, A., Girshick, R., and He, K.
\newblock Exploring randomly wired neural networks for image recognition.
\newblock In \emph{Proceedings of the IEEE/CVF International Conference on
  Computer Vision}, pp.\  1284--1293, 2019.

\bibitem[You et~al.(2020)You, Leskovec, He, and Xie]{you2020graph}
You, J., Leskovec, J., He, K., and Xie, S.
\newblock Graph structure of neural networks.
\newblock In \emph{International Conference on Machine Learning}, pp.\
  10881--10891. PMLR, 2020.

\bibitem[Zhu et~al.(2021)Zhu, Xu, Zhang, Liu, Wu, and Wang]{zhu2021deep}
Zhu, Y., Xu, W., Zhang, J., Liu, Q., Wu, S., and Wang, L.
\newblock Deep graph structure learning for robust representations: A survey,
  2021.

\end{thebibliography}
\bibliographystyle{icml2022}

\newpage
\appendix
\onecolumn

\section{Details of the Proposed Framework}
\label{algodescription}
\subsection{VGAE-based LSO}
\paragraph{Training VGAE.}
Recent research has shown that disjoint training of variational auto-encoder and Gaussian process model is less prone to overfitting comparing to joint training scheme\,\cite{siivola2021good}. In such decoupled training scenario, optimization is often found challenging due to the comparably small portion of well-performing region within the search space. Regarding that, weighted retraining\,\cite{tripp2020sampleefficient} has appeared to be a promising method which enables effective learning of feasible latent space, and a number of subsequent work has emerged based on the retraining method\,\cite{notin2021improving, verma2021uncertaintyaware, grosnit2021high}. Following such consensus, we frequently retrained VGAE with the incremented data along with the optimization process. Our training objective is as follows:
\begin{equation}
\label{trainingobj}
    L_{model} = L_{VAE} + \alpha \cdot L_{metric} + \beta \cdot L_{reg}
\end{equation}
To provide proper guidance in learning latent space, we combine deep metric learning \cite{grosnit2021high}. Specifically, log-ratio loss \cite{Kim2019DeepML} is used for $L_{metric}$, i.e.,
\begin{equation}
    L_{metric} = \left[ log \frac{\left\lVert z_{i}-z_{neg} \right\rVert_{2}}{\left\lVert z_{i}-z_{pos} \right\rVert_{2}}  - log \frac{\left\lvert f(z_{i})-f(z_{neg}) \right\rvert}{\left\lvert f(z_{i})-f(z_{pos}) \right\rvert} \right]^{2}
\end{equation}
where we sample the positive counterpart of a sample by selecting the one that has the least difference in its evaluation value and vice versa. We set the training as minimization of the objective, thus $L_{VAE}$ is the negative term of variational lower bound where the structure reconstruction term is replaced by feature reconstruction, i.e.,
\begin{equation}
    L_{VAE} = \mathcal{D}_{KL}[q(\mathbf{Z|\mathbf{X, A}})\parallel g(\mathbf{Z})] - \mathbb{E}_{q(\mathbf{Z|\mathbf{X, A}})}[g(\mathbf{X}|\mathbf{Z})]
\end{equation}
Because we have a mixture of discrete and continuous inputs, the loss term for feature reconstruction consists of mixture of regression loss and classification loss. For each case, we used Frobenius norm loss and Brier loss and the term is weighted regarding its ranking, i.e. weight $\omega \propto \frac{1}{kN + rank(\mathbf{x})},\: rank(\mathbf{x})=|\{\mathbf{x_{i}}: f(\mathbf{x}) > f(\mathbf{x}_{i}), \mathbf{x}_{i}\in\mathcal{X}\}|$. $L_{reg}$ is an orthogonal regularization term for GCN encoder\,\cite{kipf2016variational}. We use $k=10^{-2}$ for weight function hyperparameter and the objective hyperparameters are set as $\alpha=0.1$ and $\beta=0.1$.

As mentioned in \autoref{approach}, we used VGAE from \citealp{kipf2016variational} with 2-layer MLP as decoder and 1-layer MLPs for projection layers. With the training objective in \autoref{trainingobj}, the model is first warm-up trained at the beginning of the optimization and then frequently retrained with the incrementing data along with the optimization process. For warm-up training, the model is trained for 5 epochs and retrained with the frequency of 1. The pseudocode is presented in \autoref{alg:prior}.

\begin{algorithm}[h]
   \caption{Latent Space Optimization Framework (with Prior Graph)}
   \label{alg:prior}
\begin{algorithmic}
   \STATE {\bfseries Input:} initial observations $\mathcal{X}_{init}$, budget $T$, generative/inverse model $g$/$q$, surrogate model $\mathcal{M}$, prior graph $\mathcal{G}$
   \STATE Initialize $\mathcal{X}_{0} = \mathcal{X}_{init}$
   \STATE Warm-up train generative/inverse model $g$/$q$ with $\mathcal{X}_{0}$ 
   \FOR{$t=1$ {\bfseries to} $T$}
   \STATE Extract graph embeddings $Z \leftarrow q(\mathcal{X}_{t-1}, \mathcal{G})$
   \STATE Update $\mathcal{M}$ with $Z$ and $f$ and optimize the surrogate model hyperparameters
   \STATE Select next point $\mathbf{z^{t}}$; decode it to obtain candidate $\mathbf{x^{t}}$ and query evaluation $f(\mathbf{x^{t}})$
   \STATE Augment data $\mathcal{D}_{t} \leftarrow \mathcal{D}_{t-1} \cup (\mathbf{x^{t}}, f(\mathbf{x^{t}}))$
   \STATE Retrain generative/inverse model $g$/$q$ with $\mathcal{D}_{t}$
   \ENDFOR
\end{algorithmic}
\end{algorithm}
\setlength{\textfloatsep}{10pt}

\paragraph{Latent Space Optimization (LSO).}
The latent space is formed by embedded graph representation. Referred from the preceding works\,\,\cite{scarselli2009ieee, bridge2020}, additional global node is attached with incoming edges from all existing nodes, so that the learned representation of the global node is used as a graph representation. Thus, the learned embedding is aware of the structural information. The latent space is bounded by a hypercube that encompasses all embeddings with a margin of standard deviation for each dimension $m\in[1, M]$, i.e.,
\begin{equation}
    min_{\mathbf{z^{t}}\in Z^{t}}(z^{tm}) - \sigma(z^{tm}) \leq z^{m} \leq max_{\mathbf{z^{t}}\in Z^{t}}(z^{tm}) + \sigma(z^{tm})
\end{equation}
where $\mathbf{z^{t}} = [z^{t1},\dots,z^{tM}]$ for iteration $t$. We use GP-BO with Mat$\acute{e}$rn kernel as a search strategy in the latent space and the GP hyperparameters are optimized by multi-started gradient descent. For acquisition function, we used UCB with scale parameter $\kappa=2.0$.

\subsection{Nested MAB}
\paragraph{Graph Generation of BA-biased Model.} For implementation, we modified the original BA model code from networkx package\footnote[4]{\url{https://networkx.org/documentation/stable/reference/generators.html}}. As an input, we provide complete graph G with a subset of nodes that are sampled by $Agent_{\mathcal{N}}$ as described in \autoref{proposed}. Note that since the number of nodes $c$ to be centered is flexible, we put a threshold value to make the generate graph prone to be centered by the prior graph even in the case when $c=1$, i.e. single-node graph. Specifically, we set the threshold to for all experiments.

\begin{center}
\begin{tabular}{c}
\begin{lstlisting}[language=Python]
def _random_subset(seq, m, rng):
    targets = set()
    while len(targets) < m:
        x = rng.choice(seq)
        targets.add(x)
    return targets
    
def generate_graph(n, threshold, initial_graph=G):
    # List of existing nodes added proportionally to their degree
    # threshold to ensure the centeredness
    repeated_nodes = [c for c, d in G.degree() for _ in range(max(threshold, d))] 

    # Start adding the other n - m nodes.
    node_list = list(range(n))
    for node in set(repeated_nodes):
        node_list.remove(node)
        
    while len(node_list) != 0:
        node = node_list.pop(0)
        # Now choose a single node from the existing nodes
        # Pick uniformly from repeated_nodes (preferential attachment)
        targets = _random_subset(repeated_nodes, 1, seed)
        # Add edge to 1 node from the source.
        G.add_edges_from(zip([node] * 1, targets))
        # Add the node to the list which the new edge has been just connected.
        repeated_nodes.extend(targets)
        # And the new node "source" to the list.
        repeated_nodes.extend([node] * 1)

    return G

\end{lstlisting}
\end{tabular}
\end{center}

\paragraph{Graph Replacement.} The replacement mechanism would induce imbalanced training, i.e. the initial set of graphs would be trained more frequently than the replaced graphs, so we allocate separate encoders for different graphs to resolve such issue. For instance, we implement $K$ encoders each matched with $K$ number of generated graphs, therefore when graph $i\in[1,K]$ is selected, $i^{th}$ encoder is used for encoding. When the graph is replaced, corresponding encoder model parameter is re-initialized as well.

\section{Experimental Setting}
\label{expsetting}
We use open-source code implementation of 5 baseline methods\footnotemark[5]. For the hyperparameter $\lambda$ of $k_{mixture}$ kernel used in CoCaBO and CASMOPOLITAN, we follow the setting of CASMOPOLITAN by fixing $\lambda=0.5$ for all experiments. For GEBO, the dimensionality of the latent space is set to 4. The experiment results are reported by mean and standard deviation over 10 random repetitions with 40 initial points.

\footnotetext[5]{SMAC: \url{https://github.com/automl/SMAC3}, TPE: \url{https://github.com/hyperopt/hyperopt}, CoCaBO: \url{https://github.com/rubinxin/CoCaBO_code}, 
CASMOPOLITAN: \url{https://github.com/xingchenwan/Casmopolitan},
HyBO: \url{https://github.com/aryandeshwal/HyBO}
}

\section{Additional Experiment Results}
\label{appendixexp}
\begin{figure*}[ht]
\vskip 0.1in
\begin{center}
\centerline{\includegraphics[width=\textwidth]{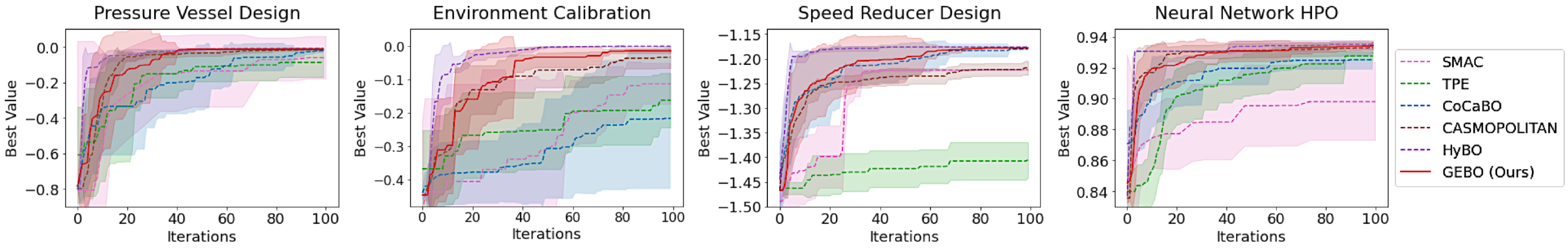}}
\caption{Comparison with the baseline models over 100 iterations.}
\label{iterexp}
\end{center}
\end{figure*}

\autoref{iterexp} shows comparison of the methods by iterations. HyBO exceeds all the method both in convergence and performance. However, as shown in \autoref{timecomparison}, computation cost needed to obtain next candidate is far more expensive comparing to other methods, hence making its practical application impossible. CASMOPOLITAN also shows competitive performance. Yet, it often underperforms comparing to GEBO due to the 
deficient consideration on pairwise interaction between variables resulted from the hand-designed kernel.

\begin{table}[h]
\caption{Computation cost in average wall-clock time (seconds) per one iteration. \label{timecomparison}}
\vskip 0.15in
\begin{center}
\begin{small}
\begin{sc}
\resizebox{0.7\textwidth}{!}{\begin{tabular}{lccccccr}
\toprule
Task & SMAC & TPE & CoCaBO & CASMOPOLITAN & HyBO & \textbf{GEBO}\\
\midrule
\makecell[l]{Environment\\Calibration} & 0.302 & 0.045 & 0.263  & 1.981 & 87.032 & \textbf{0.278}\\\midrule
\makecell[l]{Pressure Vessel\\Design}  & 0.263 & 0.057 & 0.376 & 1.781 & 82.520 & \textbf{0.425} \\\midrule
\makecell[l]{Speed Reducer\\Design}    & 1.185 & 0.054 & 0.243 & 1.818 & 74.572 & \textbf{0.426}\\\midrule
\makecell[l]{Neural Network\\HPO}    & 1.292 & 0.769 & 0.697 & 2.078 & 152.252 & \textbf{0.694}\\
\midrule
\makecell[l]{Robot Pushing\\Control}    & 1.124 & 0.123 & 0.630 & 4.569 & 134.190 & \textbf{0.546}\\
\midrule
\makecell[l]{Ackley53C}         & 1.134 & 0.340 & 2.549 & 4.583 & 489.800 & \textbf{0.890}\\
\bottomrule
\end{tabular}}
\end{sc}
\end{small}
\end{center}
\end{table}

TPE requires very less time for its computation, however, it does not take the interaction information into account, thus showing low performance either in respect of time or iteration budget. On the other hand, CoCaBO takes similar amount of time for computation comparing to GEBO, but the cost increases as the dimension of discrete subspace increases.

\begin{figure*}[h]
\vskip 0.1in
\begin{center}
\centerline{\includegraphics[width=0.6\textwidth]{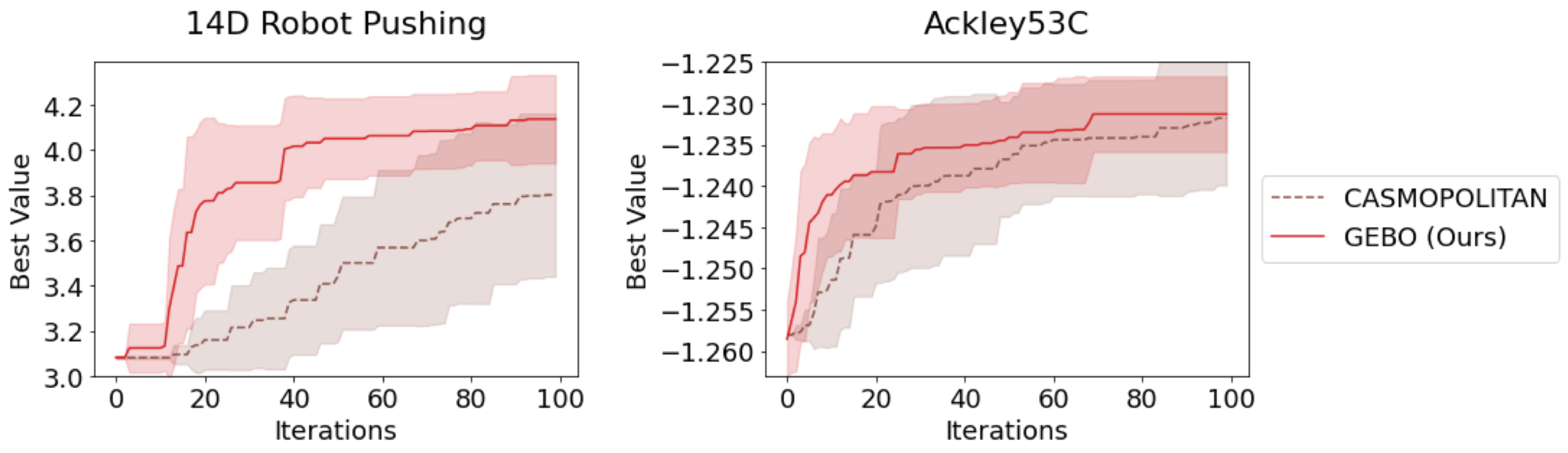}}
\caption{Comparison of GEBO and CASMOPOLITAN in high-dimensional problems.}
\label{highcomparison}
\end{center}
\vskip -0.2in
\vspace{-5pt}
\end{figure*}

We provide comparison of CASMOPOLITAN and GEBO for the two high-dimensional problems in \autoref{highcomparison}. Although CASMOPOLITAN is specifically designed to tackle high-dimensional cases, GEBO outperforms for both tasks, where we can infer that utilizing relational information is essential to achieve efficient optimization. 



\end{document}